\documentclass[twoside]{article}

\usepackage[accepted]{aistats2022}
\usepackage{bookmark}
\usepackage{mathtools}
\usepackage{mathalfa}
\usepackage{mathrsfs}
\usepackage{textcomp}
\usepackage{gensymb}
\usepackage{floatrow}
\usepackage{booktabs}
\usepackage{enumerate}
\usepackage{algorithmic}
\usepackage{algorithm}

\newcommand{\aldo}[1]{}

\usepackage{chngcntr}
\counterwithin{figure}{section}

\usepackage{amssymb,amsmath,amsthm}
\usepackage{thmtools, thm-restate}
\usepackage[ruled, vlined]{algorithm2e}

\numberwithin{equation}{section}

\input m.macros
\input m.symbols

\usepackage{tikz}
\usetikzlibrary{external}
\usetikzlibrary{positioning}
\usetikzlibrary{intersections,shapes.arrows}
\usepackage{tkz-graph}

\tikzset{
        decision/.style={circle,draw=black!60,fill=gray!10,thick, inner sep=0pt,minimum size=5mm}
        }
\tikzset{
        reward/.style={rectangle,draw=black!60,fill=blue!10,thick, inner sep=0pt,minimum size=5mm}
        }
\tikzset{
        badreward/.style={rectangle,draw=black!60,fill=red!10,thick, inner sep=0pt,minimum size=5mm}
        }

\newtheorem{lemma}{Lemma}[section]

\newtheorem{corollary}{Corollary}[section]

\usepackage[round]{natbib}

\usepackage{hyperref}
\usepackage[capitalize]{cleveref}

\begin{document}

%

%

\twocolumn[

\aistatstitle{Towards an Understanding of Default Policies\\
in Multitask Policy Optimization}

\aistatsauthor{ Ted Moskovitz \And Michael Arbel \And  Jack Parker-Holder \And Aldo Pacchiano }

\aistatsaddress{ Gatsby Unit, UCL \And  Inria \And Oxford University \And Microsoft Research } ]

\begin{abstract}
    Much of the recent success of deep reinforcement learning has been driven by regularized policy optimization (RPO) algorithms with strong performance across multiple domains. In this family of methods, agents are trained to maximize cumulative reward while penalizing deviation in behavior from some reference, or \emph{default} policy. In addition to empirical success, there is a strong theoretical foundation for understanding RPO methods applied to single tasks, with connections to natural gradient, trust region, and variational approaches. However, there is limited formal understanding of desirable properties for default policies in the \emph{multitask} setting, an increasingly important domain as the field shifts towards training more generally capable agents. Here, we take a first step towards filling this gap by formally linking the quality of the default policy to its effect on optimization. Using these results, we then derive a principled RPO algorithm for multitask learning with strong performance guarantees. 
\end{abstract}

\section{INTRODUCTION}
Appropriate regularization has been a key factor in the widespread success of policy-based deep reinforcement learning (RL) \citep{Levine:2018_rlai,Furuta:2021_rpo_overview}. The key idea underlying such \emph{regularized policy optimization} (RPO) methods is to train an agent to maximize reward while minimizing some cost which penalizes deviations from useful behavior, typically encoded as a \emph{default policy}. In addition to being easily scalable and compatible with function approximation, these methods have been shown to ameliorate the high sample complexity of deep RL methods, making them an attractive choice for high-dimensional problems \citep{dota,Espeholt:2018_impala}.

A natural question underlying this success is \emph{why} these methods are so effective. Fortunately, there is a strong foundation for the formal understanding of regularizers in the single-task setting. These methods can be seen as approximating a form of natural gradient ascent \citep{Kakade:2002,Pacchiano:2019,Moskovitz:2021}, trust region or proximal point optimization \citep{Schulman:2015,Schulman:2017}, or variational inference \citep{Levine:2018_rlai,Haarnoja:2018,Marino:2020_iterative,Abdolmaleki:2018}, and thus are well-understood by theory \citep{Agarwal:2020_pg_theory}. 

However, as interest has grown in training general agents capable of providing real world utility, there has been a shift in emphasis towards \emph{multitask} learning. Accordingly, there are a number of approaches to learning or constructing default policies for regularized policy optimization in multitask settings \citep{Galashov:2019,Teh:2017_distral,Goyal:2019,Goyal:2020,Tirumala:2020_behavior}. The basic idea is to obtain a default policy which is generally useful for some family of tasks, thus offering a form of supervision to the learning process. However, there is little theoretical understanding of how the choice of default policy affects optimization. Our goal in this paper is to take a first step towards bridging this gap, asking: 

\vspace{2mm}
\noindent\emph{\textbf{Q1:} What properties does a default policy need to have in order to improve optimization on new tasks?}
\vspace{2mm}

This is a nuanced question. The choice of penalty, structural commonalities among the tasks encountered by the agent, and even the distribution space in which the regularization is applied have dramatic effects on the resulting algorithm and the agent's performance characteristics. 

In this work, we focus on methods using the Kullback-Leibler (KL) divergence with respect to the default policy, as they are the most common in the literature. We first consider this form of regularized policy optimization applied to a single task, with the goal of understanding how the relationship between the default and optimal policies for a given problem affect optimization. We then generalize these results to the multitask setting, where we not only quantify the advantages of this family of approaches, but also identify its limitations, both fundamental and algorithm-specific. 

In the process of garnering new understanding of these algorithms, our results also imply a new framework through which to understand families of tasks. Because different algorithms are sensitive to different forms of structure, this leads to another guiding question, closely tied to the first: 

\vspace{2mm}
\noindent\emph{\textbf{Q2:} What properties does a group of tasks need to share for a given algorithm to provide a measurable benefit?}
\vspace{2mm}

It's clear that in order to be effective, any multitask learning algorithm must be applied to a task distribution with some form of structure identifiable by that algorithm: if tasks have nothing in common, no understanding gained from one task will be useful for accelerating learning on another. Algorithms may be designed to accommodate---or learn---a broader array of structures, but at increased computational costs. In high-dimensional problems, function approximation mandates new compromises. In this paper, which we view as a first step towards understanding these trade-offs, we make the following contributions: 

\begin{itemize}
    \item We show the error bound and iteration complexity for optimization using an $\alpha$-optimal default policy, where sub-optimality is measured via the distance from the optimal policy for a given task. 
    \item From these results, we derive a principled RPO algorithm for multitask learning, which we term \emph{total variation policy optimization} (TVPO). We show that popular multitask KL-based algorithms can be seen as approximations to TVPO and demonstrate the strong performance of TVPO on simple tasks.
    \item We offer novel insights on the optimization characteristics---both limitations and advantages---of common multitask RPO frameworks in the literature. 
\end{itemize}


\section{REGULARIZED POLICY OPTIMIZATION}
\paragraph{Reinforcement learning} In reinforcement learning (RL), an agent learns how to act within its environment in order to maximize its performance on a given task or tasks. We model a task as a finite \emph{Markov decision process} (MDP; \citep{Puterman:2010}) $M = (\St, \A, P, r, \gamma, \rho)$, where $\St,\A$ are finite state and action spaces, respectively, $P: \St \times \A \to \Delta(\St)$ is the state transition distribution, $r: \St \times \A \to [0,1]$ is a reward function, $\gamma\in[0,1)$ is the discount factor, and $\rho \in \Delta(\St)$ is the starting state distribution. $\Delta(\cdot)$ is used to denote the simplex over a given space. We also assume access to a restart distribution for training $\mu\in\Delta(\mathcal S)$ such that $\mu(s) > 0\ \forall s\in\St$, as is common in the literature \citep{Kakade_Langford:2002_pfd,Agarwal:2020_pg_theory}. The agent takes actions using a stationary policy $\pi: \St \to \Delta(\A)$, which, in conjunction with the transition dynamics, induces a distribution over trajectories $\tau = (s_t, a_t)_{t=0}^\infty$.

The value function $V^\pi: \St \to \reals^+$ measures the expected discounted cumuluative reward obtained by following $\pi$ from state $s$, $V^\pi(s) \coloneqq \expect[\pi]{\sum_{t=0}^\infty \gamma^t r(s_t, a_t)|s_t = s}$, where the expectation is with respect to the distribution over trajectories induced by $\pi$ in $M$. We overload notation and define $V^\pi(\rho) \coloneqq \expect[s_0\sim\rho]{V^\pi(s_0)}$ as the expected value for initial state distribution $\rho$. The action-value and advantage functions are given by
\begin{align*}
    Q^\pi(s,a) &\coloneqq \expect[\pi]{\sum_{t=0}^\infty \gamma^t r(s_t, a_t)|s_t = s, a_t=a},\\
    A^\pi(s,a) &\coloneqq Q^\pi(s,a) - V^\pi(s). 
\end{align*} 
By $d_{s_0}^{\pi}$, we denote the discounted state visitation distribution of $\pi$ with starting state distribution $\mu$, so that 
\begin{align}
\begin{split}
    d_{s_0}^{\pi}(s) &= \expect[s_0\sim\mu]{(1-\gamma)\sum_{t=0}^\infty \gamma^t \mathrm{Pr}^\pi(s_t=s|s_0)},
\end{split}
\end{align} 
where $d_\mu^{\pi}\coloneqq \expect[s_0\sim\mu]{d_{s_0}^{\pi}(s)}$.
The goal of the agent is to adapt its policy so as to maximize its value, i.e., optimize $\max_\pi V^\pi(\rho)$. We use $\pi^\star \in \argmax_\pi V^\pi(\rho)$ to denote the optimal policy and $V^\star$ and $Q^\star$ as shorthand for $V^{\pi^\star}$ and $Q^{\pi^\star}$, respectively.  

\paragraph{Policy parameterizations} In practice, this problem typically takes the form $\max_{\theta \in \Theta} V^{\pi_\theta}$, where $\{\pi_\theta | \theta\in\Theta\}$ is a class of parametric policies. In this work, we primarily consider the softmax policy class, which may be tabular or \emph{complete} (able to represent any stochastic policy), as in the case of the tabular softmax 
\begin{align} \label{eq:direct_softmax}
    \pi_\theta(a|s) = \frac{\exp(\theta_{s,a})}{\sum_{a'\in\A} \exp(\theta_{s,a'})},
\end{align}
where $\theta \in \reals^{|\St|\times|\A|}$, or \emph{restricted}, where $\pi_\theta(a|s) \propto \exp(f_\theta(s,a))$, with $f_\theta: \St \times \A \to \reals$ some parametric function class (e.g., a neural network). 

The general form of the \emph{regularized policy optimization} (RPO) objective function is given by 
\begin{align}
    \mathcal J_\lambda(\theta) \coloneqq V^{\pi_\theta}(\mu) - \lambda \Omega(\theta), 
\end{align}
where $\Omega$ is some convex regularization functional. Gradient ascent updates proceed according to 
\begin{align} \label{eq:grad_ascent}
    \theta^{(t+1)} = \theta^{(t)} + \eta \grad_\theta \mathcal J_\lambda(\theta^{(t)}).
\end{align}
For simplicity of notation, from this point forward, for iterative algorithms which obtain successive estimates of parameters $\theta^{(t)}$, we denote the associated policy and value functions as $\pi^{(t)}$ and $V^{(t)}$, respectively. The choice of $\Omega$ plays a signficant role in algorithm design and practice, as we discuss below. It's also important to note that the error bounds and convergence rates we derive are based on the basic policy gradient framework in Appendix \cref{alg:po}, in which update \cref{eq:grad_ascent} is applied across a batch after every $B$ trajectories $\{\tau_b\}_{b=1}^B$ are sampled from the environment. Therefore, the iteration complexities below are proportional to the associated sample complexity. 

\section{RELATED WORK}
\paragraph{Single-task learning} Most theoretical \citep{Agarwal:2020_pg_theory,Grill:2020} and empirical \citep{Schulman:2015,Schulman:2017,Abdolmaleki:2018,Pacchiano:2019} literature has focused on the use of RPO in a single-task setting, i.e., applied to a single MDP. These methods often place a soft or hard constraint on the Kullback-Leibler (KL) divergence between the updated policy at each time step and the current policy, maximizing an objective of the form 
\begin{align}
\begin{split}
\mathcal J_\lambda(\pi_p, \pi_q) = \sum_{t=0}^\infty \mathbb E_{s_t\sim d_\mu^{\pi_{q}}} &\mathbb E_{a_t\sim\pi_q(\cdot|s_t)} \big[\mathcal G(s_t,a_t)  \\ &- \lambda \kl(\pi_p(\cdot|s_t), \pi_q(\cdot|s_t))\big],
\end{split}
\end{align}
where $\mathcal G: \mathcal S \times \mathcal A \to \reals$ is typically the $Q$- or advantage function and $\pi_q, \pi_p \in \{\pi_\theta, \pi_0\}$ \citep{Furuta:2021_rpo_overview}. At each update, then, the idea is to maximize reward while minimizing the regularization cost. From a theoretical perspective, such methods can often be framed as a form of approximate variational inference, with either learned \citep{Abdolmaleki:2018,Song:2019_vmpo_forward,Peng:2020_awr_forward,Nair:2021_awac_reverse,Peters:2010_reps} or fixed \citep{Todorov:2007_klcontrol,Toussaint:2006_infmdp,Rawlik:2013_kl,Tishby:2016_glearning} $\pi_0$. 
When $\pi_0 \approx \pi_\theta$, we can also understand such approaches as approximating the natural policy gradient \citep{Kakade:2002}, which is known to accelerate convergence \citep{Agarwal:2020_pg_theory}. Similarly, regularizing the objective using the Wasserstein distance \citep{Pacchiano:2019} rather than the KL divergence produces updates which approximate those of the Wasserstein natural policy gradient \citep{Moskovitz:2021}. Other approaches can be understood as trust region or proximal point methods \citep{Schulman:2015,Schulman:2017,Touati:2020_newtrpo}, or even model-based approaches \citep{Grill:2020}. It's also important to note the special case of entropy regularization, where $\Omega(\theta) = -\mathbb E_{s\sim\Unif_\St} \mathsf H[\pi_\theta(\cdot|s)] = \mathbb E_{s\sim\Unif_\St} \kl(\pi_\theta(\cdot|s), \Unif_\A)$ (where $\Unif_\mathcal X$ denotes the uniform distribution over a space $\mathcal X$) which is perhaps the most common form of RPO \citep{Haarnoja:2018,Levine:2018_rlai,Mnih:2016_a3c,Williams:1991_entropy,Schulman:2018_entropy} and has been shown to aid optimization by encouraging exploration and smoothing the objective function landscape \citep{Ahmed:2019_entropy}.

\paragraph{Multitask learning}
Less common in the literature are policy regularizers designed explicitly for multitask settings. In many multitask RL algorithms which apply RPO, shared task structure is leveraged in other forms (e.g., importance weighting), and the regularizer itself doesn't reflect shared information \citep{Espeholt:2018_impala,Riedmiller:2018_playing}. However, in cases where the penalty is designed for multitask learning, the policy is penalized for deviating from a more general \emph{task-agnostic} default policy meant to encode behavior which is generally useful for the \emph{family} of tasks at hand. The use of such a behavioral default is intuitive: by distilling the common structure of the tasks the agent encounters into behaviors which have shown themselves to be useful, optimization on new tasks can be improved with the help of prior knowledge. For example, some approaches \cite{Goyal:2019,Goyal:2020} construct a default policy by marginalizing over goals $g$ for a set of goal-conditioned policies $\pi_0(a|s) = \sum_g P(g)\pi_\theta(a|s,g)$. Such partitioning of the input into goal-dependent and goal-agnostic features can be used to create structured internal representations via an information bottleck \citep{Tishby:2000}, shown empirically to improve generalization. In other multitask RPO algorithms, the default policies are derived from a Bayesian framework which views $\pi_0$ as a prior \citep{Wilson:2007_bayesian,Osband:2020_TS}. Still other methods learn $\pi_0$ online through distillation \citep{Hinton:2015_distillation} by minimizing $\kl(\pi_0, \pi)$ with respect to $\pi_0$ \citep{Galashov:2019,Teh:2017_distral}. When $\pi_0$ is preserved across tasks but $\pi_\theta$ is re-initialized, $\pi_0$ learns the average behavior across task-specific policies. However, to our knowledge, there has been no investigation of the formal optimization properties of explicitly multitask approaches, and basic questions remain unanswered. 


\section{A BASIC THEORY FOR DEFAULT POLICIES}
At an intuitive level, the question we'd like to explore is: \emph{What properties does a default policy need in order to improve optimization?} By ``improve'' we refer either to a reduction in the error at convergence with respect to the optimal value function or a reduction in the number of updates required to reach a given error threshold. To begin, we consider perhaps the simplest default: the uniform policy. The proofs for this section are provided in \cref{sec:appendix_singletask}.

\subsection{Log-Barrier Regularization}
For now, we'll restrict ourselves to the direct softmax parameterization (\cref{eq:direct_softmax})
with access to exact gradients. Our default is a uniform policy over actions, i.e.: $\pi_0(a|s) = \Unif_\A$, resulting in the objective 
\begin{align}
\begin{split}
    \Ju(\theta) &\coloneqq V^{\pi_\theta}(\mu) - \lambda \expect[s\sim\Unif_\St]{\kl(\Unif_\A, \pi_\theta(\cdot|s))} \\
        &\equiv V^{\pi_\theta}(\mu) + \frac{\lambda}{|\St||\A|} \sum_{s,a} \log\pitheta,
\end{split}
\end{align}
where we have dropped terms that are constant with respect to $\theta$. 
Importantly, it's known that even this default policy has beneficial effects on optimization by erecting a log-barrier against low values of $\pi_\theta(a|s)$. This barrier prevents gradients from quickly dropping to zero due to exponential scaling, leading to a polynomial convergence rate\footnote{It remains an open question whether entropy regularization, which is gentler in penalizing low probabilities, produces a polynomial convergence rate.}. We now briefly restate convergence error and iteration complexity results for this case, first presented by \cite{Agarwal:2020_pg_theory} (Theorem 5.1 and Corollary 5.1, respectively): 
\begin{lemma}[Error bound for log-barrier regularization] \label{thm:log-barrier_err}
    Suppose $\theta$ is such that $\| \nabla_\theta \Ju(\theta) \|_2 \leq \epsilon_{\mathrm{opt}}$, with $\epsilon_{\mathrm{opt}} \leq \frac{\lambda}{2 |\St||\A|} $. Then for all start state distributions $\rho$,
    \begin{align*} 
        V^{\pi_\theta}(\rho) &\geq V^\star(\rho) - \frac{2\lambda}{1 - \gamma}\Dinf.
    \end{align*}
\end{lemma}
We briefly comment on the term $\Dinf$ (in which the division refers to component-wise division), known as the \emph{distribution mismatch coefficient}, which roughly quantifies the difficulty of the exploration problem faced by the optimization algorithm. We note that $\mu$ is the starting distribution used for training/optimization, while the ultimate goal is to perform well on the target starting state distribution $\rho$. The iteration complexity is given below.
\begin{lemma}[Iteration complexity for log-barrier regularization] \label{thm:log-barrier_iter}
    Let $\beta_\lambda\coloneqq \frac{8\gamma}{(1 - \gamma)^3} + \frac{2\lambda}{|\St|}$. Starting from any initial $\theta^{(0)}$, consider the updates \cref{eq:grad_ascent} with $\lambda = \frac{\eps(1 - \gamma)}{2\Dinf}$ and $\eta = 1/\beta_\lambda$. Then for all starting state distribution $\rho$, we have  
    \begin{align*}
        &\min_{t\leq T} \{V^\star(\rho) - V^{(t)}(\rho)\} \leq \eps \quad 
        \\ &\text{whenever} \quad T \geq \frac{320|\St|^2|\A|^2}{(1 - \gamma)^6\eps^2}\Dinf^2.
    \end{align*}
\end{lemma}
These results will act as useful reference points for the following investigation. At a minimum, we'd like a default policy to provide guarantees that are at least as good as those of log-barrier regularization.

\subsection{Regularization With An \texorpdfstring{$\vec\alpha$}{alpha}-Optimal Policy}
To understand the properties required of the default policy, we place an upper bound on the suboptimality of $\pi_0$ via the TV distance. For each $s\in\St$, let 
\begin{align}
    d_{\mathrm{TV}}(\pi^\star(\cdot|s), \pi_0(\cdot|s)) \leq \alpha(s)
\end{align}
Our regularized objective is 
\begin{align}
\begin{split} \label{eq:jalpha}
    \Jz(\theta) &= V^{\pi_\theta}(\mu) - \lambda\expect[s\sim\Unif_\mathcal S]{\kl(\pi_0(\cdot|s), \pi_\theta(\cdot|s))} \\
    &\equiv V^{\pi_\theta}(\mu) + \frac{\lambda}{|\St|} \sum_{s,a} \pi_0(a|s) \log \pitheta 
\end{split}
\end{align}
for starting state distribution $\mu\in\Delta(\mathcal S)$. We then have
\begin{align} \label{eq:jalpha_grad}
\begin{split}
    \Partial{\Jz(\theta)}{\theta_{s,a}} = 
     \frac{1}{1 - \gamma} d^{\pi_\theta}_\mu(s) &\pi_\theta(a|s) A^{\pi_\theta}(s,a) \\ &+ \frac{\lambda}{|\St|} (\pi_0(a|s) - \pi_\theta(a|s)).
\end{split}
\end{align}

Our first result presents the error bound for first-order stationary points of the $\pi_0$-regularized objective.
\begin{restatable}[Error bound for $\alpha(s)$-optimal $\pi_0$]{lemma}{errtight} \label{thm:alpha_reg_tight}
    Suppose $\theta$ is such that $\| \nabla \Jz(\theta) \|_2 \leq \epsilon_{\mathrm{opt}}$. Then we have that for all starting distributions $\rho$:
  \begin{align*}
    &V^{\pi_\theta}(\rho) \geq V^*(\rho) - \min\Biggr\{\frac{1}{1-\gamma} \times
     \\ &\expect[s\sim\Unif_{\St}]{\frac{\epsilon_{\mathrm{opt}} |\mathcal{S}|}{\max\left\{1 - \alpha(s) - \frac{\epsilon_{\mathrm{opt}}|\mathcal{S}|}{\lambda}, 0\right\}}  + \lambda \alpha(s)} \Dinf,\\
    &\frac{|\A| - 1}{(1-\gamma)^2}\left(\expect[s\sim\mu]{\alpha(s)} \left\| \frac{d_\rho^{\pi_\theta}}{\mu} \right\|_\infty  +  \frac{\epsilon_{\mathrm{opt}} |\St|}{\lambda} \right) 
    \Biggr\}
\end{align*}
\end{restatable}

The $\min\{\cdot\}$ operation above reflects the fact that the value of $\lambda$ effectively determines whether reward-maximization or the regularization dominates the optimization of \cref{eq:jalpha}. Note that a similar effect also applies to log-barrier regularization, but the ``high'' $\lambda$ setting is excluded in that instance because as $\lambda \to \infty$, $\pitheta \to \Unif_\A$. In this case, however, as $\alpha\to0$, a high value of $\lambda$ might be preferable, as it would amount to doing supervised learning with respect to a (nearly) optimal policy. When the reward-maximization dominates, we can see that the error bound becomes vacuous as $\alpha(s)$ approaches $\alpha^- \coloneqq 1 - \eps_{\mathrm{opt}}|\St|/\lambda$ from below. In other words, as $\alpha$ approaches this point, the error can grow arbitrarily high. 

In the KL-minimizing case, we can see that as the policy error $\alpha\to0$, the value gap is given by $\frac{\epsilon_{\mathrm{opt}} |\St|(|\A| - 1)}{\lambda(1 - \gamma)^2}$. Intuitively, then, as the default policy moves closer to $\pi^\star$, we can drive the value error to zero as $\lambda\to\infty$. Interestingly, we can also see that as the distribution mismatch $\|d_\rho^{\pi^\star}/\mu\|_\infty\to0$, the influence of the policy distance $\alpha$ diminishes and the error can again be driven to zero by increasing $\lambda$. We leave a more detailed discussion of the impact of the distribution mismatch coefficient to future work. Note that in most practical cases, neither $\alpha$ nor $\|d_\rho^{\pi^\star}/\mu\|_\infty$ will be low enough to achieve a lower error via KL minimization alone. We will therefore focus on the reward-maximizing case ($\lambda < 1$) for the majority of our further analysis. 

Before considering iteration complexity however, it's also helpful to note that \cref{thm:alpha_reg_tight} generalizes \cref{thm:log-barrier_err} given the same upper-bound on $\eps_\mathrm{opt}$ as \cite{Agarwal:2020_pg_theory}.
\begin{restatable}{corollary}{err} \label{thm:alpha_reg_sd}
    Suppose $\theta$ is such that $\| \nabla \Jz(\theta) \|_{\infty} \leq \epsilon_{\mathrm{opt}}$, with $\epsilon_{\mathrm{opt}} \leq \frac{\lambda}{2 |\St||\A|} $ and $\lambda < 1$. Then we have that for all states $s\in\St$,
    \begin{align*} 
        V^{\pi_\theta}(\rho) \geq V^\star(\rho) - 
            \frac{\expect[s\sim\Unif_{\St}]{\kappa_\A^\alpha(s)} \lambda}{1-\gamma} \Dinf
    \end{align*}
    where $\kappaA = \frac{2|\A|(1 - \alpha(s)) }{2|\A|(1 - \alpha(s)) - 1}$.
\end{restatable}

We can see that in this case, the coefficient $\kappaA$ takes on key importance. In particular, we can see that the error-bound becomes vacuous as $\alpha(s)$ approaches $\alpha^- =  1 - 1/(2|\A|)$ from below.  The error bound is improved with respect to log-barrier regularization when the coefficient $\kappaA < 2$, which occurs for $\alpha(s) < 1 - 1/|\A|$. Note that this is the TV distance between the uniform policy and a deterministic optimal policy.
\begin{figure*}[!t]
    \centering
    \includegraphics[width=0.99\textwidth]{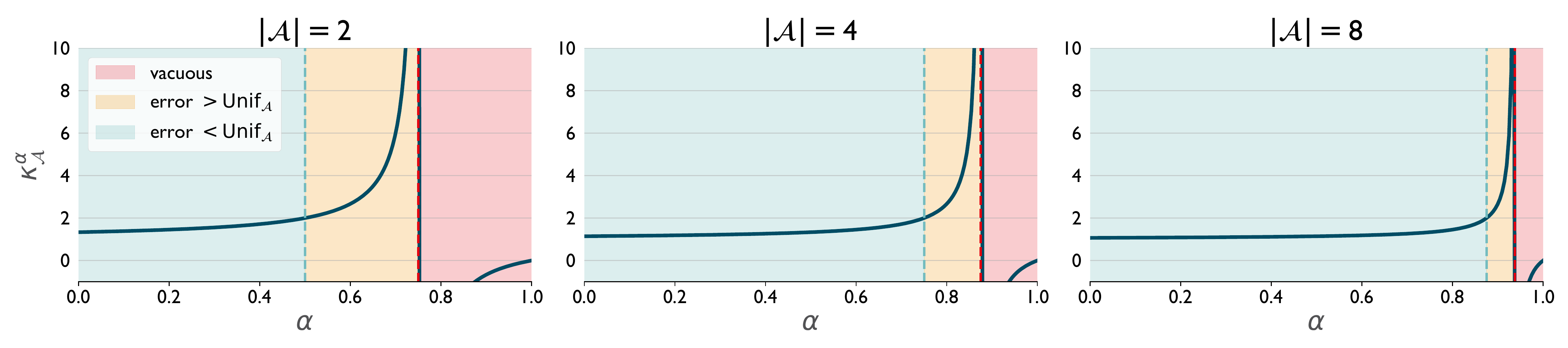}
    \caption{As $|\A|$ grows, regularizing using $\pi_0$ with larger $\tv(\pi^\star(\cdot|s), \pi_0(\cdot|s))$ will converge to a lower error than log-barrier regularization. In other words, there is a more forgiving margin of error for the default policy.}
    \label{fig:kappa}
\end{figure*}
These relationships are visualized in \cref{fig:kappa}. We can see that the range of values over which $\alpha$-optimal regularization will result in lower error than log-barrier regularization grows as the size of the action space increases. This may have implications for the use of a uniform default policy in continuous action spaces, which we leave to future work.


We can then combine this result with standard results for the convergence of gradient ascent to first order stationary points to obtain the iteration complexity for convergence. First, however, we require an upper bound on the smoothness of $\Jz$ as defined in \cref{eq:jalpha}.
\begin{restatable}[Smoothness of $\Jz$]{lemma}{smoothness} \label{thm:smoothness_jalpha} For the softmax parameterization, we have that 
    \begin{align*}
        ||\grad_\theta \Jz(\theta) - \grad_\theta \Jz(\theta')||_2 \leq \beta_\lambda ||\theta - \theta'||_2
    \end{align*}
where  $\beta_\lambda = \frac{8}{(1 - \gamma)^3} + \frac{2\lambda}{|\St|}$. 
\end{restatable}

We can now bound the iteration complexity.

\begin{restatable}[Iteration complexity for $\Jz$]{lemma}{itercomplexity} \label{thm:jalpha_iters}
    Let $\rho$ be a starting state distribution. Following \cref{thm:smoothness_jalpha}, let $\beta_\lambda= \frac{8\gamma}{(1-\gamma)^3} + \frac{2\lambda}{|\St|}$. From any initial $\theta^{(0)}$ and following \cref{eq:grad_ascent} with $\eta = 1/\beta_\lambda$ and 
    \begin{align*}
        \lambda = \frac{\eps(1 - \gamma)}{\expect[s\sim\Unif_{\St}]{\kappa_\A^\alpha(s)} \Dinf} < 1,
    \end{align*}
    we have   
    \begin{align*}
        &\min_{t\leq T} \{V^\star(\rho) - V^{(t)}(\rho)\} \leq \eps \\ &\mathrm{whenever} \quad T \geq  \frac{80 \expect[s\sim\Unif_{\St}]{\kappaA}^2|\St|^2|\A|^2}{(1 - \gamma)^6\eps^2} \Dinf^2.
    \end{align*}
\end{restatable}

It is also natural to consider the case in which $\pi_0$ is used as an initialization for $\pi_\theta$. 

\begin{restatable}[]{corollary}{itercompinit}\label{thm:jalpha_iters_pi0}
Given the same assumptions as \cref{thm:jalpha_iters}, if the initial policy is chosen to be $\pi_0$, i.e., $\pi_{\theta^{(0)}} = \pi_0$ where $\pi_0(\cdot|s)$ is $\alpha(s)$-optimal with respect to $\pi^\star(\cdot|s)$ $\forall s$, then  
 \begin{align*}
    &\min_{t\leq T} \{V^\star(\rho) - V^{(t)}(\rho)\} \leq \eps \\ &\mathrm{whenever} \  
        T \geq \frac{320\vert\mathcal{A} \vert^2 \vert \mathcal{S}\vert^2}{\epsilon^2(1-\gamma)^7} \left\Vert\frac{d_{\rho}^{\pi^{\star}}}{\mu} \right\Vert_{\infty}^2\left\Vert\frac{1}{\mu} \right\Vert_{\infty}
        \mathbb{E}_{s\sim \mu}\left[\alpha(s)\right].
\end{align*}
\end{restatable}
In the case of random initialization, note that when $\alpha(s) = \alpha = 1 - 1/|\A|$, $\mathbb E \kappaA = 2$, recovering the iteration complexity for log-barrier regularization, as expected. We also see that as the error $\alpha$ moves higher or lower than $1 - 1/|\A|$, the iteration complexity grows or shrinks quadratically. Therefore, a default policy within this range will not only linearly reduce the error at convergence, but will also quadratically increase the rate at which that error is reached. 
When the initial policy is $\pi_0$, the iteration complexity depends on the factor $\mathbb{E}_{s\sim \Unif_{\mathcal{S}}}[\alpha(s)]$. Hence, for good initialization, $\alpha$ is small, resulting in fewer iterations. The natural question, then, is how to find such a default policy, with high probability, for some family of tasks. 

\section{MULTITASK LEARNING}
The results above provide guidance for the construction of default policies in the multitask setting. The key insight is that if the optimal policies for the tasks drawn from a given task distribution have commonalities, the agent can use the optimal policies it learns from previous tasks to construct a useful $\pi_0$. More precisely, consider a distribution $\mathcal P_\mathcal M$ over a family of tasks $\mathcal M\coloneqq \{M_k\}$. (The simplest example of such a distribution is a categorical distribution over a discrete set of tasks, although continuous distributions over MDPs are possible.) We assume only that the tasks have shared state and action spaces $\mathcal S$ and $\A$, and we denote their optimal deterministic policies by $\{\pi^\star_k\}$. 
We assume that the other task components (reward function, transition distribution, etc.) are independent. Then by \cref{thm:alpha_reg_sd} and \cref{thm:jalpha_iters}, if the TV barycenter at a given state $s$, given by 
\begin{align} \label{eq:pi0_opt}
    \pi_0(\cdot|s) &= \argmin_\pi \expect[M_k\sim\mathcal P_\mathcal M]{\tv(\pi_k^\star(\cdot|s), \pi(\cdot|s))} 
\end{align}
is such that $\mathbb E [\tv(\pi_k^\star(\cdot|s), \pi_0(\cdot|s))] < 1 - 1/|\A|$, then regularizing with $\pi_0$ will, in expectation, result in faster convergence and lower error than using a uniform distribution. Crucially, when there is a lack of shared structure, which in this particular approach is manifested as a lack of agreement among optimal policies, $ \pi_0(\cdot|s)$ collapses to $\Unif_{\mathcal{A}}$. Therefore, in the worst case, regularizing with $ \pi_0(\cdot|s)$ can do no worse than log-barrier regularization, which already enjoys polynomial iteration complexity.

With deterministic optimal policies, 
the following gives a convenient expression for the TV barycenter:
\begin{restatable}[TV barycenter]{lemma}{tvbarycenter} \label{thm:tv_barycenter}
    Let $\mathcal P_{\mathcal M}$ be a distribution over tasks $\mathcal M = \{M_k\}$, each with a deterministic optimal policy $\pi_k^\star: \mathcal S \to \A$. Define 
    \begin{align}
        \xi(s,a) \coloneqq \expect[M_k\sim\mathcal P_\mathcal M]{\mathbf 1(\pi_k^\star(s) = a)}.
    \end{align}
    as the average optimal action. Then, the TV barycenter $\pi_0(\cdot|s)$ defined in \cref{eq:pi0_opt} is given by a greedy policy over $\xi$, i.e., $\pi_0(a|s) = \delta(a \in \argmax_{a'\in\A} \xi(s,a'))$, where $\delta(\cdot)$ is the Dirac delta distribution. 
\end{restatable}
The proof, along with the rest of the proofs for this section, is provided in \cref{sec:appendix_multitask}.
Interestingly, this result also holds for the KL barycenter, which we show in Appendix \cref{thm:kl_bary}. Because the average optimal action $\xi$ is closely related to a recently-proposed computational model of \emph{habit formation} in cognitive psychology \citep{Miller:2019}, from now on we refer to it as the \emph{habit function} for task family $\mathcal M$. 
When the agent has observed $K$ tasks sampled from $\mathcal P_\mathcal M$,  $\xi$ is approximated by the sample average $\hat\xi(s,a) = \frac{1}{K} \sum_{k=1}^K \mathbf 1(\pi_k^\star(s) = a)$ provided that the optimal policies $\pi_k^\star$ are available. 
In practice, however, the agent only has access to an approximation $\tilde{\pi}_k$ of $\pi_k^\star$ obtained, for instance, through the use of a learning algorithm $A$, such as Appendix \cref{alg:rpo}. Hence, $\hat\xi(s,a) $ is instead given by $\hat\xi(s,a) = \frac{1}{K} \sum_{k=1}^K \mathbf 1(\tilde{\pi}_k(s) = a)$ which induces an approximate barycenter $\hat{\pi}_0$ by taking the greedy policy over $\hat{\xi}$. The following result provides the iteration complexity for the multitask setting when using $\hat{\pi}_0$ as the default policy.
\begin{restatable}[Multitask iteration complexity]{lemma}{multitaskiters}\label{lem:avg_complexity}
    Let $M_k \sim \mathcal P_\mathcal M$ and denote by $\pi_k^{\star}: \St \to \A$ its  optimal policy. Denote by $T_k$ the number of iterations to reach $\epsilon$-error for $M_k$ in the sense that:
    \begin{align*}
        \min_{t\leq T_k} \{V^{\pi_k^\star}(\rho) &- V^{(t)}(\rho)\} \leq \eps.
    \end{align*}
    Set $\lambda, \beta_\lambda$, and $\eta$ as in \cref{thm:jalpha_iters}. From any initial $\theta^{(0)}$, and following \cref{eq:grad_ascent}, $\expect[M_k\sim\mathcal P_\mathcal M]{T_k}$ satisfies:
    \begin{align*}
        \expect[M_k\sim\mathcal P_\mathcal M]{T_k} \geq  \frac{80\vert\mathcal{A} \vert^2 \vert \mathcal{S}\vert^2}{\epsilon^2(1-\gamma)^6} \mathbb E_{\substack{M_k\sim\mathcal P_\mathcal M\\
        s\sim\Unif_\St}}\left[\kappa_\A^{\alpha_k}(s) \left\| \frac{d_{\rho}^{\pi_k^*} }{\mu}  \right \|_\infty^2\right],
    \end{align*}
    where $\alpha_k(s) \coloneqq \tv(\pi_k^\star(\cdot|s), \hat{\pi}_0(\cdot|s))$. If $\hat{\pi}_0$ is also used for initialization, then $\expect[M_k\sim\mathcal P_\mathcal M]{T_k}$ satisfies:
    \begin{align*}
        \expect[M_k\sim\mathcal P_\mathcal M]{T_k} \geq  \frac{320\vert\mathcal{A} \vert^2 \vert \mathcal{S}\vert^2}{\epsilon^2(1-\gamma)^7}\left\Vert\frac{1}{\mu} \right\Vert_{\infty}^3 \mathbb E_{\substack{M_k\sim\mathcal P_\mathcal M\\
        s\sim\mu}}\left[ \alpha_k(s) \right],
    \end{align*}
\end{restatable}
\cref{lem:avg_complexity} characterizes the average iteration complexity over tasks when using $\hat{\pi}_0$ as a default policy. In particular, when the learning algorithm is also initialized with $\hat{\pi}_0$, we obtain that the average number of iterations to reach $\epsilon$ accuracy is proportional to the expected  TV distance of $\hat{\pi}_0$ to the optimal policies $\pi_k^{\star}$ for tasks $\{M_k\}\sim\mathcal P_{\mathcal M}$. We expect this distance to approach  $\mathbb{E}\left[\tv(\pi_0(\cdot|s),\pi_k^{\star}(\cdot|s))\right]$ as the number of tasks increases and $\tilde{\pi}_k$ become more accurate. Note that even in this case, the regularization is \emph{still} required to assure polynomial convergence. To provide a precise quantification,  we let $\tilde{\pi}_k(\cdot|s)$ be, on average, $\zeta(s)$-optimal in state $s$ across tasks $\{M_k\}$, i.e.  $\mathbb{E}_{M_k\sim \mathcal{M}}\left[\tv(\tilde\pi_k(\cdot|s),\pi_k^{\star}(\cdot|s))\right]\leq \zeta(s)$ for some $\zeta(s)\in [0, 1]$. The following lemma quantifies how  close $\hat{\pi}_0$ grows to the TV barycenter of $\{\pi_k^{\star}\}_{k=1}^K$ as $K\to\infty$: 

\begin{restatable}[Barycenter concentration]{lemma}{barycentercon}
Let $\delta$ be  $0<\delta<1$. Then with  probability higher than $1-\delta$, for all $s\in\St$, it holds that:
\begin{align*}
    \vert \mathbb{E}_{M_k\sim \mathcal P_\mathcal{M}}[ \tv(&\pi^{\star}_k(\cdot|s), \hat{\pi}_0(\cdot|s)) - \tv(\pi^{\star}_k(\cdot|s), \pi_0(\cdot|s))]\vert \\ &\leq 2\zeta(s) + \sqrt{\frac{2\log(\frac{2}{\delta})}{K}} + 2C\sqrt{\frac{\vert\mathcal{A} \vert}{K}},
\end{align*}
for some constant $C$ that depends only on $\vert \mathcal{A}\vert$.
\end{restatable}
In other words, in order to produce a default policy which improves over log-barrier regularization as $K\to\infty$, the margin of error for the trained policies is half that which is required for the default policy. 

In practice, due to the epistemic uncertainty about the task family early in training (in other words, when only a few tasks have been sampled), regularizing using $\hat{\pi}_0$ risks misleading $\pi_\theta$ by placing all of the default policy's mass on a sub-optimal action. We can therefore define $\hat{\pi}_0$ using a softmax $\hat{\pi}_0(a|s) \propto \exp( \hat{\xi}(s,a)/\beta(k))$ with some temperature parameter $\beta(k)$ which tends to zero as the number of observed tasks $k$ approaches infinity. Therefore, $\pi_0$ converges to the optimal default policy in the limit. 
This suggests the simple approach to multitask RPO presented in \cref{alg:ok}, which we call \emph{total variation policy optimization} (TVPO). 
\begin{algorithm}[t]
    \caption{TV Policy Optimization (TVPO)}\label{alg:ok}
	\begin{algorithmic}[1]
    \STATE \textbf{Input} Task set $\mathcal M$, policy class $\Theta$, fixed-$\pi_0$ RPO algorithm $A(M, \Theta, \pi_0, \lambda)$, as in Appendix \cref{alg:rpo}  
    \STATE initialize $\pi_0(\cdot|s) = \xi^{(0)}(s,\cdot) = \Unif_\A$ $\forall s\in\St$
    \FOR{iteration $k=1,2,...$}
    \STATE Sample a task $M^{(k)} \sim \mathcal P_\mathcal M$
    \STATE Solve the task: $\tilde\theta^{(k)} = A(M_k, \Theta, \pi_0^{(k-1)}, \lambda)$
    \STATE Set $\tilde{\pi}_k \gets \pi_{\tilde{\theta}^{(k)}}$.
    \STATE Update habit moving average $\forall (s,a) \in \St \times \A$:
        \begin{align*}
            \xi^{(k)}(s,a) \gets  \frac{k-1}{k}&\xi^{(k-1)}(s,a) 
            \\&+ \frac{1}{k}\mathbf{1}\left(a\in\argmax_{a'}\tilde{\pi}_{k}(a'|s)\right) 
        \end{align*}
    \STATE Update default policy $\forall (s,a) \in \St \times \A$:
        \begin{align*}
            \pi_0^{(k)}(a|s) \propto \exp(\xi^{(k)}(s,a)/\beta(k)) 
        \end{align*}
    \ENDFOR
    \end{algorithmic}
\end{algorithm}
Note that if $\mathcal P_\mathcal M$ is non-stationary, the moving average in Line 8 can be changed to an exponentially weighted moving average to place more emphasis on recent tasks.

\section{UNDERSTANDING THE LITERATURE}
As stated previously, many approaches to multitask RPO in the literature learn a default policy $\pi_0(a|s;\phi)$ parameterized by $\phi$ via gradient descent on the KL divergence \citep{Galashov:2019,Teh:2017_distral}, e.g., via 
\begin{align} \label{eq:kl_obj}
    \phi = \argmin_{\phi'} \expect[s\sim\Unif_\St]{\kl(\pi_\theta(\cdot|s), \pi_0(\cdot|s;\phi))}.
\end{align}
The idea is that by updating $\phi$ across multiple tasks, $\pi_0$ will acquire the average behaviors of the goal-directed policies $\pi_\theta$. This objective can be seen as an approximation of \cref{eq:pi0_opt} in which we can view the use of the KL as a relaxation of the TV distance:
\begin{align*}
    \pi_0(\cdot|s) &= \argmin_\pi \expect[M_k \sim \mathcal P_{\mathcal M}]{\tv(\pi^\star_k(\cdot|s), \pi(\cdot|s))} \\
    &\leq  \argmin_\pi \expect[M_k \sim \mathcal P_{\mathcal M}]{\tv(\pi^\star_k(\cdot|s), \pi(\cdot|s))^2} \\
    &\leq \argmin_\pi \expect[M_k \sim \mathcal P_{\mathcal M}]{\kl(\pi^\star_k(\cdot|s), \pi(\cdot|s))},
\end{align*}
where the first inequality is due to Jensen's inequality and the second is due to \cite{Pollard:2000_kl} and where $\pi_\theta(\cdot|s) \approx \pi^\star(\cdot|s)$. The use of the KL is natural due to its easy computation and differentiability, however, the last approximation is crucial. Distilling $\pi_0$ from $\pi_\theta$ from the outset of each task implicitly assumes that $\pi_\theta \approx \pi^\star$ even early in training. This is a source of suboptimality, as we discuss in \cref{sec:expts}.

\section{EXPERIMENTS} \label{sec:expts}
\begin{figure}[!t]
       \centering\begin{tikzpicture}
            \tikzstyle{VertexStyle}=[decision]
            \Vertex[L=$s_1$,x=2,y=0]{X}

            \Vertex[L=$s_2$,x=1.3,y=-0.7]{A}
            \Vertex[L=$s_4$,x=0.6,y=-1.4]{Y}
            \tikzstyle{VertexStyle}=[badreward]
            \Vertex[L=$0$,x=0.2,y=-2.1]{Y1}
            \Vertex[L=$0$,x=1.0,y=-2.1]{Y2}
            \Vertex[L=$0$,x=1.6,y=-1.4]{Z}
            
            \tikzstyle{VertexStyle}=[decision]
            \Vertex[L=$s_3$,x=2.7,y=-0.7]{B}
            \tikzstyle{VertexStyle}=[badreward]
            \Vertex[L=$0$,x=2.3,y=-1.4]{C}            
            \tikzstyle{VertexStyle}=[decision]
            \Vertex[L=$s_5$,x=3.4,y=-1.4]{D} 
            
            \tikzstyle{VertexStyle}=[decision]
            \Vertex[L=$s_6$,x=2.7,y=-2.1]{E}
            \tikzstyle{VertexStyle}=[badreward]
            \Vertex[L=$0$,x=3.9,y=-2.1]{F}
            
            \tikzstyle{VertexStyle}=[decision]
            \Vertex[L=$s_7$,x=2.0,y=-2.8]{G}
            \tikzstyle{VertexStyle}=[badreward]
            \Vertex[L=$0$,x=3.1,y=-2.8]{H}
            
            \tikzstyle{VertexStyle}=[decision]
            \Vertex[L=$s_8$,x=1.3,y=-3.5]{I}
            \Vertex[L=$s_9$,x=2.7,y=-3.5]{J}
            \tikzstyle{VertexStyle}=[reward]
            \Vertex[L=?,x=0.9,y=-4.2]{I1}
            \Vertex[L=?,x=1.7,y=-4.2]{I2}
            \Vertex[L=?,x=2.3,y=-4.2]{J1}
            \Vertex[L=?,x=3.1,y=-4.2]{J2}            
            
            \Edges(A, X,B)
            \Edges(Y1, Y, Y2)
            \Edges(Y, A, Z)
            \Edges(C, B, D)
            \Edges(E, D, F)
            \Edges(G, E, H)
            \Edges(I, G, J)
            \Edges(I1, I, I2)
            \Edges(J1, J, J2)

        \end{tikzpicture}
        \caption{A tree environment. Each task in the family randomly distributes rewards among leaves marked with a `?'. All other states result in zero reward.}
        \label{fig:tree}
\end{figure}
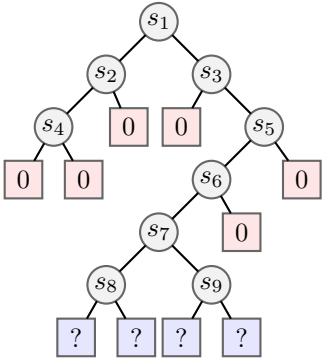

We now study the implications of these ideas in a simple empirical setting: a family of tasks whose state space follows the tree structure shown in \cref{fig:tree}. In these tasks, the agent starts at the root $s_1$ and at each timestep chooses whether to proceed down its left subtree or right subtree ($|\A| = 2$). The episode ends when the agent reaches a leaf node. In this setup, there is zero reward in all states other than the leaf nodes marked with a `?', for which one or more are randomly assigned a reward of 1 for each draw from the task distribution. To encourage sparsity, the number of rewards is drawn from a geometric distribution with success parameter $p=0.5$. 
One training run consisted of five rounds of randomly sampling a task and solving it. Despite the simplicity of this environment, we found that it could prove surprisingly difficult for many algorithms to solve consistently. As can be seen in \cref{fig:tree}, the key structural consistency in this task is that every optimal policy makes the same choices in states $\{s_1, s_3, s_5, s_6\}$, with necessary exploration limited to the lower subtree rooted at $s_7$. 

\begin{figure*}[!t]
    \centering
    \includegraphics[width=0.99\textwidth]{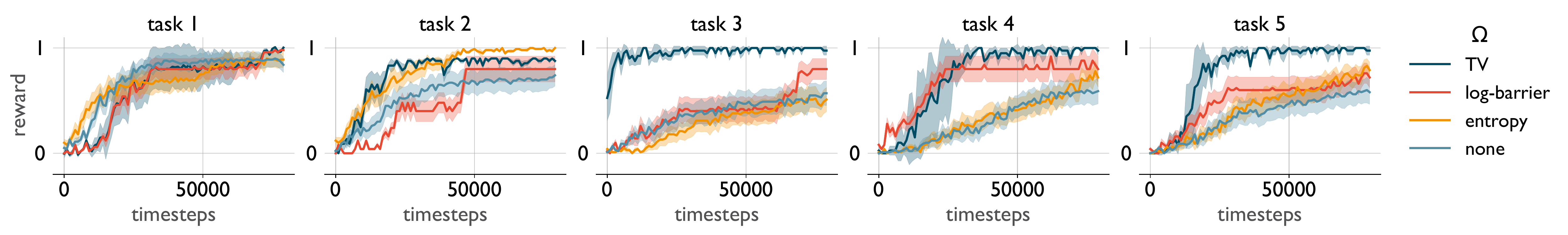}
    \caption{Fixed $\pi_0$ baselines. Results are averaged over 20 seeds, with the shaded region denoting one standard deviation.}
    \label{fig:tabular_trees_fixed}
\end{figure*}

\begin{figure*}[!t]
    \centering
    \includegraphics[width=0.99\textwidth]{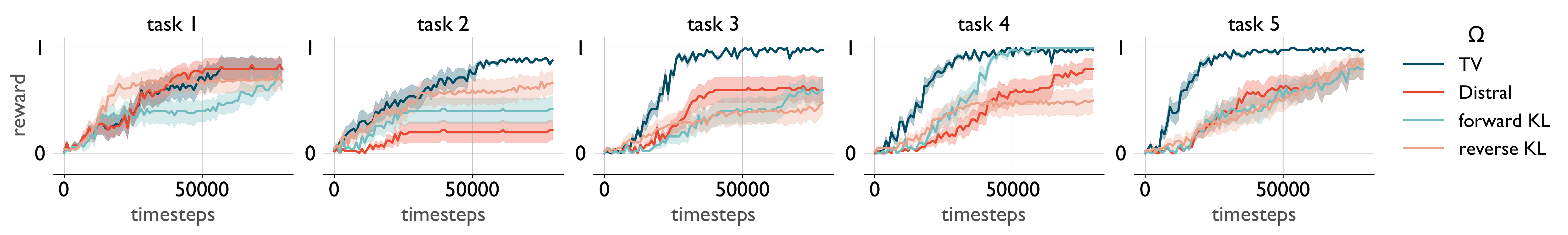}
    \caption{Learned $\pi_0$ baselines. Results are averaged over 20 seeds, with the shaded region denoting one standard deviation.}
    \label{fig:tabular_trees_learned}
\end{figure*}

For comparison, we selected RPO approaches with both \emph{fixed} default policies (\textsc{log-barrier}, \textsc{entropy}, and \textsc{none}) and \emph{learned} default policies: \textsc{Distral} ($-\kl(\pi_\theta,\pi_0) + \mathsf H[\pi_\theta]$; \citep{Teh:2017_distral}), \textsc{forward KL} ($-\kl(\pi_0,\pi_\theta)$), and \textsc{reverse KL} ($-\kl(\pi_\theta,\pi_0)$). To make the problem more challenging for the learned default policies, the reward distribution was made sparser by setting $p=0.7$. Each approach was applied over 20 random seeds, with results plotted in \cref{fig:tabular_trees_fixed} (fixed $\pi_0$) and \cref{fig:tabular_trees_learned} (learned $\pi_0$). Hyperparameters were kept constant across methods (further experimental details can be found in \cref{sec:exp_details}). We see that \textsc{TVPO} most consistently solves the tasks. This is not surprising, as $\expect[M_k\sim\mathcal P_\mathcal M]{\alpha_k(s)} = 0$ for all states en route to the rewarded leaves until $s_7$. Thus, $\hat\pi_0(\cdot|s)\to\pi^\star_k(s)$ quickly for these states as the number of tasks grows. This dramatically reduces the size of the exploration problem for TVPO, confining it to the subtree rooted at $s_7$.

\begin{figure}[!t]
    \centering
    \includegraphics[width=0.99\textwidth]{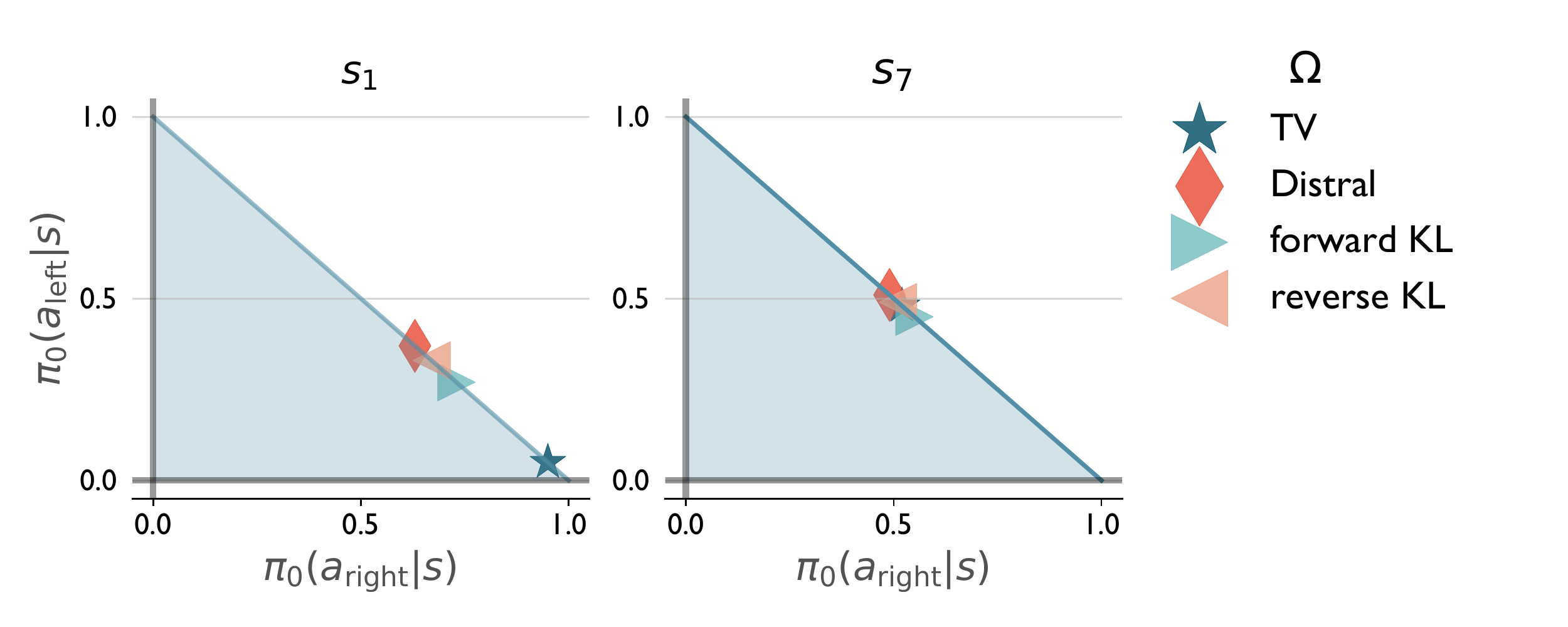}
    \caption{Learned default policies in states $s_1$ and $s_7$ after five tasks. In the simplex for $s_7$, the marker for \textsc{TVPO} is behind the markers for the other methods.}
    \label{fig:simplex5}
\end{figure}

To gain a better understanding of the results and the learned default policies, we plotted the average default policies for each method on the 2-simplex for states $s_1$ and $s_7$ in \cref{fig:simplex5}. For all tasks in the family, the optimal policy goes right in $s_1$, while, on average, reward could be located in either subtree rooted at $s_7$. This is reflected in the default policies, which prefer right in $s_1$ and are close to uniform in $s_7$. There is a notable difference, however, in that the KL-, gradient-based methods are much less deterministic in $s_1$. The critical difference is that the KL-based methods are trained online via distillation from suboptimal $\pi_\theta\not\approx \pi^\star$. Early in training, $\pi_\theta$ is inconsistent across tasks and runs, resulting in a more uniform target for $\pi_0$. This delays its convergence across tasks to the shared TV/KL barycenter. To test this effect empirically, we repeated the same experiment with \textsc{reverse KL} but started training $\pi_0$ progressively later within each task. 
\begin{figure}[H]
    \centering
    \includegraphics[width=0.85\textwidth]{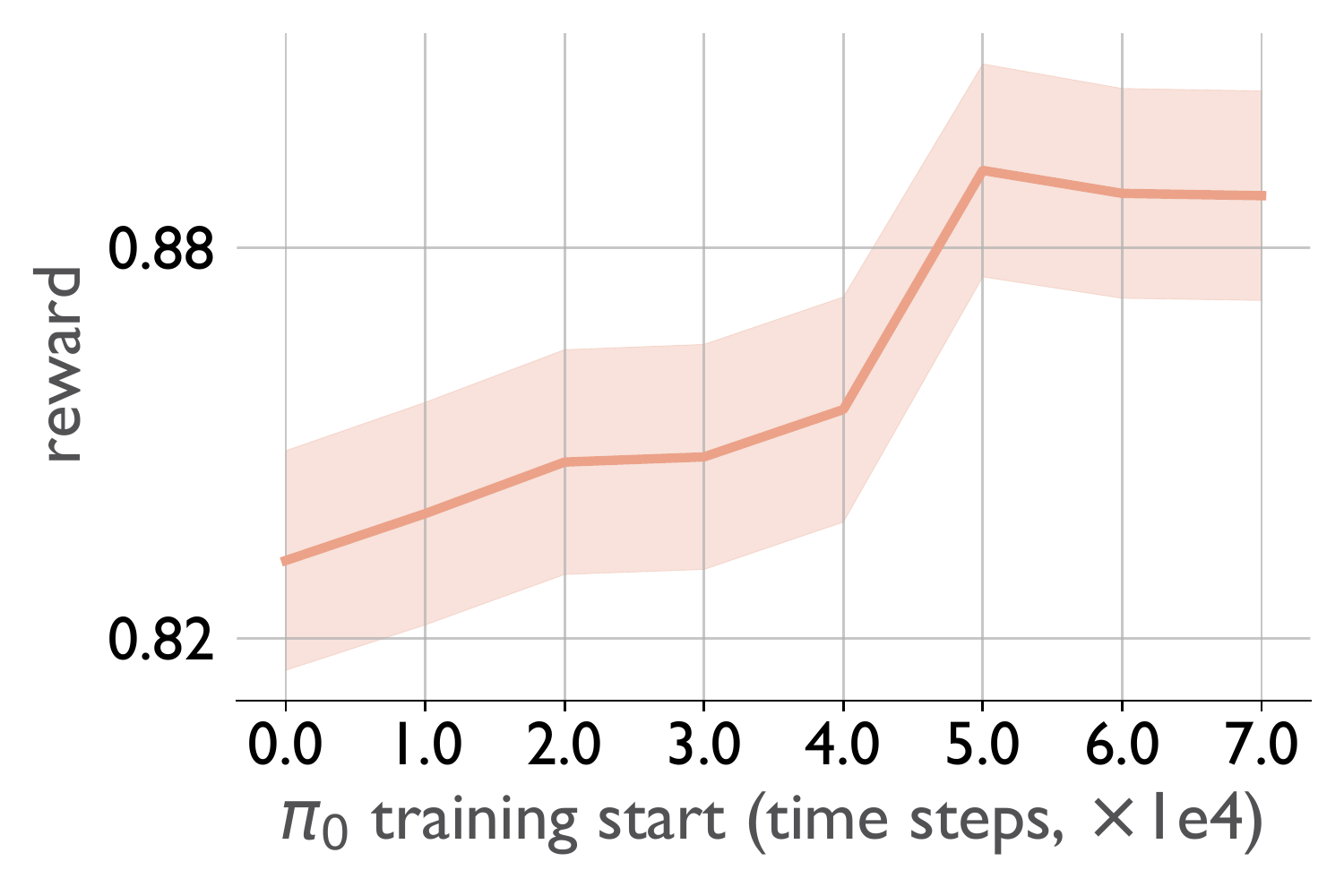}
    \caption{Delayed training of $\pi_0$ improves performance.}
    \label{fig:delay}
\end{figure}
\noindent\cref{fig:delay} depicts the average final reward across tasks for different time steps at which the default policy began training. Note, however, that $\pi_0$ is still used to regularize $\pi_\theta$, it just isn't updated based on $\pi_\theta$ until $\pi_\theta$ is a reasonable approximation of $\pi^\star$. We can see that, as predicted, delaying training within each task improves performance. There is a slight drop in performance if $\pi_0$ does not have a sufficient number of updates at the end of training.

\section{DISCUSSION} \label{sec:conclusion}
In this work, we introduce novel, more general bounds on the error and iteration complexity of $\kl$-regularized policy optimization. We then show how these bounds apply to the multitask setting, showing the first formal results for a popular class of algorithms and deriving a novel multitask RPO algorithm with formal guarantees. We then demonstrate the implications of our findings in a simple experimental setting. Taken together, we believe our results provide preliminary answers to our guiding questions for KL-regularized RPO:

\vspace{2mm}
\noindent\emph{\textbf{A1:} In order to provide benefit, a default policy must in expectation be at least as close to the optimal policy as the uniform policy.}

\noindent\emph{\textbf{A2:} For KL-regularized RPO to provide a measurable benefit, a group of tasks must induce optimal policies which agree over some portion of the state space.}
\vspace{2mm}

There are several important ramifications for future work. First, these results imply an algorithm-dependent definition of task families, such that a group of tasks can be considered a \emph{family} for a given algorithm if that algorithm can leverage their shared properties to improve optimization. For RPO algorithms, then, the choice of divergence measure, default policy, and distribution space implicitly determines task groupings. As an example, the particular class of algorithm we investigate here is sensitive to state-dependent similarities in the space of optimal policies for a group of tasks. There are a multitude of other forms of shared structure which alternative approaches can leverge, however, such as consistent transition dynamics \citep{Barreto:2020_fastgpi, Moskovitz:FR_2021} or even structure in an abstract \emph{behavioral} space \citep{Pacchiano:2019,Moskovitz:2021,Agarwal:2021}. From \cref{fig:kappa}, it is important to observe that the performance gains relative to log-barrier regularization are, ultimately, relatively small. It may be necessary to develop algorithms with stronger assumptions about the task family and/or sensitivity to a different form of structure in order to drive further improvement. Conducting an effective taxonomy of algorithms and associated task families will be crucial for the development of practical real-world agents.

We also believe this work provides a formal framework for settings where forward transfer is possible during lifelong learning scenarios with multiple interrelated tasks \citep{GradientEpisodicMemory}. 
While we tested these ideas in a toy setting, the underlying theory has implications for state-of-the-art deep RL methods. When state and action spaces grow large, however, $\pi_0$ is necessarily represented by a restricted policy class. Both TVPO and the learned $\pi_0$ baseline methods can be scaled to this domain, with TVPO's $\pi_0$ being trained online to predict the next action taken by $\pi_\theta$. One useful lesson which equally applies to KL-based methods, however, is that it's preferable from an optimization standpoint to distill $\pi_0$ from $\pi_\theta$ only late in training when $\pi_\theta\approx\pi^\star$. Given the promise of this general class of methods, we hope that the insight garnered by these results will help propel the field towards more robust and general algorithms.

\paragraph{Acknowledgments}
We'd like to thank Maneesh Sahani, Matt Botvinick, and Peter Orbanz for helpful comments and suggestions. 

\newpage
\bibliographystyle{unsrtnat} 
\bibliography{rpo}

\appendix

\onecolumn

\section{Appendix Overview}
 \cref{sec:algs} contains policy gradients pseudocode, \cref{sec:appendix_singletask} contains proofs for the single-task results (including additional results for state-dependent $\lambda$ and $\eps$), \cref{sec:appendix_multitask} contains proofs for the multitask RPO results, and \cref{sec:exp_details} contains experimental details.

\section{Generic Policy Optimization Algorithms} \label{sec:algs}
\begin{algorithm}[H]
    \caption{Generic policy gradient algorithm}\label{alg:po}
	\begin{algorithmic}[1]
    \STATE \textbf{Input} MDP $M$, policy class $\Theta$ 
    \STATE initialize $\theta^{(0)}\in\Theta$
    \FOR{iteration $k=0,1,2,\dots$}
    \STATE sample a trajectory:
        \begin{align*}
            \tau = (s_0, a_0, s_1, \dots) \sim \mathrm{Pr}^{\pi_{\theta^{(k)}}}_\mu (\cdot) = \mu(s_0) \prod_{t=0}^{\infty} P(s_{t+1}|s_t, a_t) \pi_{\theta^{(k)}}(a_t|s_t)
        \end{align*}
    \STATE update parameters:
        \begin{align*}
            \theta^{(k+1)} = \theta^{(k)} + \eta \widehat{\grad V^{\pi_\theta}}(\mu)
        \end{align*}
        where 
        \begin{align*}
            \widehat{\grad V^{\pi_\theta}}(\mu) = \sum_{t=0}^\infty \gamma^t \widehat{Q^{\pi_\theta}}(s_t, a_t) \grad \log \pi_\theta(a_t|s_t), \qquad 
            \widehat{Q^{\pi_\theta}}(s_t, a_t) = \sum_{t'=t}^\infty \gamma^{t'-t}r(s_{t'}, a_{t'})
        \end{align*}
    \ENDFOR
    \end{algorithmic}
\end{algorithm}

\begin{algorithm}[H]
    \caption{Regularized policy gradient algorithm}\label{alg:rpo}
	\begin{algorithmic}[1]
    \STATE \textbf{Input} MDP $M$, policy class $\Theta$, regularization strength $\lambda$, default policy $\pi_0$
    \STATE initialize $\theta^{(0)}\in\Theta$
    \FOR{iteration $k=0,1,2,\dots,K$}
    \STATE sample a trajectory:
        \begin{align*}
            \tau = (s_0, a_0, s_1, \dots) \sim \mathrm{Pr}^{\pi_{\theta^{(k)}}}_\mu (\cdot) = \mu(s_0) \prod_{t=0}^{\infty} P(s_{t+1}|s_t, a_t) \pi_{\theta^{(k)}}(a_t|s_t)
        \end{align*}
    \STATE update parameters:
        \begin{align*}
            \theta^{(k+1)} = \theta^{(k)} + \eta \widehat{\grad_{\theta^{(k)}} \mathcal J_\lambda}(\theta^{(k)})
        \end{align*}
        where 
        \begin{align*}
            \widehat{\grad_\theta \mathcal J_\lambda}(\theta) &= \widehat{\grad_\theta V^{\pi_\theta}}(\mu) - \lambda \grad_\theta \Omega(\pi_0, \pi_\theta)
        \end{align*}
        and $\widehat{\grad_\theta V^{\pi_\theta}}(\mu)$ is as in \cref{alg:po}. 
    \ENDFOR
    \STATE \textbf{return} $\theta^{(K)}$
    \end{algorithmic}
\end{algorithm}

\section{Single-task results} \label{sec:appendix_singletask}

We now consider the error bound for $\pi_0$ such that $d_{\mathrm{TV}}(\pi^*(\cdot|s), \pi_0(\cdot|s)) \leq \alpha(s)$ $\forall s\in\St$.

\errtight*

\begin{proof}
Let's assume that $\pi^*$ is a deterministic policy. By \cite{Puterman:2010} such an optimal policy always exists for an MDP. We'll use the notation $a^*(s)$ to denote the optimal action at state $s$. This, combined with the assumption that $d_{\mathrm{TV}}(\pi^*(\cdot|s),\pi_0(\cdot|s)) \leq \alpha(s)$ for all $s\in\St$, tells us that $\pi_0(a^*(s)|s) \geq \pi^*(a^*(s)|s) - \alpha(s) = 1 - \alpha(s)$. Similarly, for $a\neq a^*(s)$, $\pi_0(a|s) \leq \alpha(s)$. Using this, we can start by showing that whenever $A^{\pi_\theta}(s, a^*(s)) \geq 0$ we can lower bound $\pi_\theta(a^*(s) | s)$ for all $s$.

            The gradient norm assumption $ \| \nabla \Jz(\theta) \|_\infty \leq \epsilon_{\mathrm{opt}}$ implies that for all $s,a$:
            \begin{equation*}
                \epsilon_{\mathrm{opt}} \geq \Partial{\Jz(\theta)}{\theta_{s,a}} = \frac{1}{1 - \gamma} d_\mu^{\pi_\theta}(s) \pitheta A^{\pi_\theta}(s,a) + \frac{\lambda}{|\St|} (\piz - \pitheta)
            \end{equation*}
    
            In particular for all $s$,
            \begin{align}
            \begin{split} \label{eq:grad}
                \epsilon_{\mathrm{opt}} &\geq \Partial{\Jz(\theta)}{\theta_{s,a^*(s)}} \stackrel{(i)}{\geq} \frac{1}{1 - \gamma} d_\mu^{\pi_\theta}(s) \pi_\theta(a^*(s)|s) A^{\pi_\theta}(s,a^*(s)) + \frac{\lambda}{|\St|} (\pi^*(a^*(s)|s) - \alpha(s) - \pi_\theta(a^*(s)|s)) \\
                &= \frac{1}{1 - \gamma} d_\mu^{\pi_\theta}(s) \pi_\theta(a^*(s)|s) A^{\pi_\theta}(s,a^*(s)) + \frac{\lambda}{|\St|} (1 - \alpha(s) - \pi_\theta(a^*(s)|s))
            \end{split}
            \end{align}

            And therefore if $A^{\pi_\theta}(s, a^*(s)) \geq 0$,
            \begin{align*}
                \epsilon_{\mathrm{opt}} &\geq  \frac{\lambda}{|\St|} (1 - \alpha(s) - \pi_\theta(a^*(s)|s))
            \end{align*}
            Thus,
            \begin{equation}\label{eq:policies_closeness_alpha_sd}
                \pi_\theta(a^*(s)|s) \geq 1 - \alpha(s) - \frac{\epsilon_{\mathrm{opt}}|\St|}{\lambda} .
            \end{equation}
            
            We then have 
            \begin{align*}
                A^{\pi_\theta}(s, a^*(s)) &\leq \frac{1-\gamma}{d_\mu^{\pi_\theta}(s)}\left(\frac{1}{\pi_\theta(a^*(s)|s)} \Partial{\Jz(\theta)}{\theta_{s,a}} - \frac{\lambda}{|\St|} \frac{1}{\pi_\theta(a^*(s)|s)} \left(1 - \alpha(s) - \pi_\theta(a^*(s)|s)\right) \right) \\ 
                &= \frac{1-\gamma}{d_\mu^{\pi_\theta}(s)}\left(\frac{1}{\pi_\theta(a^*(s)|s)} \Partial{\Jz(\theta)}{\theta_{s,a}} + \frac{\lambda}{|\St|}\left(1 - \frac{1 - \alpha(s)}{\pi_\theta(a^*(s)|s)} \right) \right) \\
                &\stackrel{(i)}{\leq} \frac{1}{\mu(s)}\left(\frac{1}{\max\left\{1 - \alpha(s) - \frac{\epsilon_{\mathrm{opt}}|\mathcal{S}|}{\lambda}, 0\right\}} \cdot \epsilon_{\mathrm{opt}} + \frac{\lambda}{|\St|}\left( 1 - (1 - \alpha(s))\right) \right) \\
                &\leq \frac{1}{\mu(s)}\left(\frac{1}{\max\left\{(1 - \alpha(s)) - \frac{\epsilon_{\mathrm{opt}}|\mathcal{S}|}{\lambda}, 0\right\}} \cdot \epsilon_{\mathrm{opt}} + \frac{\lambda}{|\St|} \alpha(s)\right)
            \end{align*}
            where $(i)$ follows because $d_{\mu}^{\pi_\theta}(s)\geq (1-\gamma)\mu(s)$, $\frac{\partial \mathcal{J}_\lambda^{\alpha}(\theta)}{\partial \theta_{s,a}}\leq \epsilon_{\mathrm{opt}}$ and $\max( 1 - \alpha(s) - \frac{\epsilon_{\mathrm{opt}}|\St|}{\lambda} , 0)\leq \pi_\theta(a^*(s) | s) \leq 1$ .
            Then applying the performance difference lemma \citep{Kakade_Langford:2002_pfd} gives
            \begin{align*}
                         V^*(\rho) - V^{\pi_\theta}(\rho) &= \frac{1}{1-\gamma} \sum_{s,a} d_{\rho}^{\pi^*}(s) \pi^*(a|s) A^{\pi_\theta}(s,a) \\
                &= \frac{1}{1-\gamma} \sum_{s} d_{\rho}^{\pi^*}(s)  A^{\pi_\theta}(s,a^*(s)) \\
                &\leq \frac{1}{1-\gamma} \sum_{s} d_{\rho}^{\pi^*}(s)  A^{\pi_\theta}(s,a^*(s)) \mathbf{1}( A^{\pi_\theta}(s,a^*(s)) \geq 0 ) \\
                &\leq \frac{1}{1-\gamma}  \sum_{s} \frac{d_{\rho}^{\pi^*}(s) }{\mu(s)} \left( \frac{1}{\left\{1 - \alpha(s) - \frac{\epsilon_{\mathrm{opt}}|\mathcal{S}|}{\lambda}, 0\right\}} \cdot \epsilon_{\mathrm{opt}} + \frac{\lambda}{|\St|} \alpha(s) \right)\mathbf{1}( A^{\pi_\theta}(s,a^*(s)) \geq 0 )  \\ 
                &\leq \frac{1}{1-\gamma} \expect[s\sim\mathrm{Unif}_{\St}]{\frac{\epsilon_{\mathrm{opt}} |\mathcal{S}|}{\left\{1 - \alpha(s) - \frac{\epsilon_{\mathrm{opt}}|\mathcal{S}|}{\lambda}, 0\right\}}  + \lambda \alpha(s)} \Dinf.
         \end{align*} 
         Now let's relate the values of $\pi^*$ and $\pi_\theta$. We will again apply the performance difference lemma, this time in the other direction:
     \begin{align*}
      V^{\pi_\theta}(\rho) -V^*(\rho) &= \frac{1}{1-\gamma} \sum_{s,a} d_{\rho}^{\pi_\theta}(s) \pi_\theta(a|s) A^{\pi^*}(s,a) \\
      &\eqtext{(1)}  \frac{1}{1-\gamma} \sum_s \left( \sum_{a\neq a^*(s)} d_{\rho}^{\pi_\theta}(s) \pi_\theta(a|s) A^{\pi^*}(s,a)   \right) \\
      &\stackrel{(2)}{\geq}  \frac{-1}{1-\gamma} \sum_s d_{\rho}^{\pi_\theta}(s) \left(\alpha(s) + \frac{\epsilon_{\mathrm{opt}}|\St|}{\lambda}\right) \frac{|\A| - 1}{1-\gamma} \\
      &= -\frac{|\A| - 1}{1-\gamma} \left(\sum_s \frac{d_{\rho}^{\pi_\theta}(s)}{1 - \gamma}\alpha(s) + \frac{\epsilon_{\mathrm{opt}}|\St|}{\lambda}\sum_s \frac{d_{\rho}^{\pi_\theta}(s)}{1 - \gamma}\right) \quad (\text{b/c } \sum_s d^{\pi_\theta}_\rho(s) = 1)\\ 
      &= -\frac{|\A| - 1}{1-\gamma} \left(\sum_s \frac{d_{\rho}^{\pi_\theta}(s)}{1 - \gamma}\alpha(s) \right) - \frac{(|\A| - 1)\epsilon_{\mathrm{opt}} |\St|}{\lambda (1-\gamma)^2} \\
      &= -\frac{|\A| - 1}{(1-\gamma)^2} \left( \sum_s   d_{\rho}^{\pi_\theta}(s) \alpha(s) \right) -  \frac{(|\A| - 1)\epsilon_{\mathrm{opt}} |\St|}{\lambda (1-\gamma)^2} \\
      &=   -\frac{|\A| - 1}{(1-\gamma)^2} \left( \sum_s   \frac{d_{\rho}^{\pi_\theta}(s) }{\mu(s)} \mu(s)\alpha(s) \right) -  \frac{(|\A| - 1)\epsilon_{\mathrm{opt}} |\St|}{\lambda (1-\gamma)^2} \\
      &\geq  -\frac{|\A| - 1}{(1-\gamma)^2}\left\| \frac{d_{\rho}^{\pi_\theta} }{\mu} \right\|_\infty \left( \sum_s    \mu(s)\alpha(s) \right) -  \frac{(|\A| - 1)\epsilon_{\mathrm{opt}} |\St|}{\lambda (1-\gamma)^2} 
 \end{align*}
where (1) is due to the fact that $A^{\pi^*}(s,a^*(s)) = 0$, and (2) is due to the fact that $A^{\pi^*}(s,a)$ for $a\neq a^*$ is lower-bounded by $-1/(1-\gamma)$ and \cref{eq:policies_closeness_alpha_sd}.
Therefore,
\begin{equation*}
    V^{\pi_\theta}(\rho) + \frac{|\A| - 1}{(1-\gamma)^2}\left(\expect[s\sim\mu]{\alpha(s)} \Dinf  +  \frac{\epsilon_{\mathrm{opt}} |\St|}{\lambda} \right)  \geq V^*(\rho)  .
\end{equation*}
This completes the proof.
\end{proof}

We now present a comparatively looser bound which applies the same upper bound on the norm of the gradient used by \cite{Agarwal:2020_pg_theory}.
\err*
\begin{proof}
    The proof proceeds as in \cref{thm:alpha_reg_tight}, except that we use the upper bound on $\epsilon_{\mathrm{opt}}$ in \cref{eq:policies_closeness_alpha_sd} to get 
            \begin{equation}\label{eq:policies_closeness_alpha_sd_loost}
                \pi_\theta(a^*(s)|s) \geq 1 - \alpha(s) - \frac{\epsilon_{\mathrm{opt}}|\St|}{\lambda} \geq 1 - \alpha(s) - \frac{1}{2|\A|} = \frac{2|\A|(1 - \alpha(s)) - 1}{2|\A|}
            \end{equation}
            In this case we can upper bound $A^{\pi_\theta}(s, a^*(s))$. From \cref{eq:grad} inequality $(i)$, we have 
            \begin{align*}
                A^{\pi_\theta}(s, a^*(s)) &\leq \frac{1-\gamma}{d_\mu^{\pi_\theta}(s)}\left(\frac{1}{\pi_\theta(a^*(s)|s)} \Partial{\Jz(\theta)}{\theta_{s,a}} - \frac{\lambda}{|\St|} \frac{1}{\pi_\theta(a^*(s)|s)} \left(1 - \alpha(s) - \pi_\theta(a^*(s)|s)\right) \right) \\ 
                &= \frac{1-\gamma}{d_\mu^{\pi_\theta}(s)}\left(\frac{1}{\pi_\theta(a^*(s)|s)} \Partial{\Jz(\theta)}{\theta_{s,a}} + \frac{\lambda}{|\St|}\left(1 - \frac{1 - \alpha(s)}{\pi_\theta(a^*(s)|s)} \right) \right) \\
                &\stackrel{(i)}{\leq} \frac{1}{\mu(s)}\left(\frac{1}{(1 - \alpha(s)) - \frac{\epsilon_{\mathrm{opt}}|\mathcal{S}|}{\lambda}} \cdot \epsilon_{\mathrm{opt}} + \frac{\lambda}{|\St|}\left( 1 - (1 - \alpha(s))\right) \right) \\
                &\leq \frac{1}{\mu(s)}\left(\frac{1}{(1 - \alpha(s)) - \frac{\epsilon_{\mathrm{opt}}|\mathcal{S}|}{\lambda}} \cdot \epsilon_{\mathrm{opt}} + \frac{\lambda}{|\St|} \alpha(s)\right)  \\
                    &\stackrel{(ii)}{\leq} \frac{1}{\mu(s)}\left(\frac{2|\A|}{\left(2|\A|(1 - \alpha(s)) - 1\right)} \frac{\lambda}{2|\St||\A|} + \frac{\lambda}{|\St|}\alpha(s)\right) \\
                    &= \frac{\lambda}{|\St|\mu(s)}\left(\frac{1}{2|\A|(1 - \alpha(s)) - 1} + \underbrace{\alpha(s)}_{\leq 1} \right) \\
                    &\leq \frac{\lambda}{|\St|\mu(s)} \left( \underbrace{\frac{2|\A|(1 - \alpha(s)) }{2|\A|(1 - \alpha(s)) - 1}}_{\coloneqq \kappa_\A^\alpha(s)} \right)
            \end{align*}
            Where $(i)$ follows because $d_{\mu}^{\pi_\theta}(s)\geq (1-\gamma)\mu(s)$, $\frac{\partial \mathcal{J}_\lambda^{\alpha}(\theta)}{\partial \theta_{s,a}}\leq \epsilon_{\mathrm{opt}}$ and $\max( 1 - \alpha(s) - \frac{\epsilon_{\mathrm{opt}}|\St|}{\lambda} , 0)\leq \pi_\theta(a^*(s) | s) \leq 1$ . $(ii)$ is obtained by plugging in the upper bound on $\eps_{\mathrm{opt}}$.

            We now make use of the performance difference lemma:
         \begin{align}
             V^*(\rho) - V^{\pi_\theta}(\rho) &= \frac{1}{1-\gamma} \sum_{s,a} d_{\rho}^{\pi^*}(s) \pi^*(a|s) A^{\pi_\theta}(s,a) \\
                &= \frac{1}{1-\gamma} \sum_{s} d_{\rho}^{\pi^*}(s)  A^{\pi_\theta}(s,a^*(s)) \\
                &\leq \frac{1}{1-\gamma} \sum_{s} d_{\rho}^{\pi^*}(s)  A^{\pi_\theta}(s,a^*(s)) \mathbf{1}( A^{\pi_\theta}(s,a^*(s)) \geq 0 ) \\
                &\leq \frac{\lambda}{(1-\gamma)|\St|} \sum_s \kappa_\A^\alpha(s) \frac{d_{\rho}^{\pi^*}(s) }{\mu(s)}  \mathbf{1}( A^{\pi_\theta}(s,a^*(s)) \geq 0 ) \\
                &\leq \frac{\lambda}{(1-\gamma)} \expect[s\sim\mathrm{Unif}_{\St}]{\kappa_\A^\alpha(s)} \Dinf
         \end{align}

         This completes the proof. 

\end{proof}


We can bound the smoothness of the objective as follows. 
\smoothness*
\begin{proof}
    We can first bound the smoothness of $V^{\pi_\theta}(\mu)$ using Lemma D.4 from \cite{Agarwal:2020_pg_theory}. We get 
    \begin{align*}
        ||\grad_\theta V^{\pi_\theta}(\mu) - \grad_\theta V^{\pi_{\theta'}}(\mu)||_2 \leq \beta ||\theta - \theta'||_2
    \end{align*}
    for 
    \begin{align*}
        \beta = \frac{8}{(1 - \gamma)^3}.
    \end{align*}
    We now need to bound the smoothness of the regularizer $\frac{\lambda}{|\St|}\Omega(\theta)$ where  
    \begin{align*}
        \Omega(\theta) = \sum_{s,a} \pi_0(a|s) \log \pitheta.
    \end{align*}
    Using that $\Partial{}{\theta_{s',a'}} \log \pitheta = \mathbf{1}(s=s')[\mathbf{1}(a=a') - \pi_\theta(a'|s)]$ for the softmax parameterization, we get 
    \begin{align*}
        \grad_{\theta_s} \Omega(\theta) &= \pi_0(\cdot|s) - \pi_\theta(\cdot|s), \\
        \grad_{\theta_s}^2 \Omega(\theta) &= -\mathrm{diag}(\pi_\theta(\cdot|s)) + \pi_\theta(\cdot) \pi_\theta(\cdot|s)\tr.
    \end{align*}
    The remainder of the proof follows directly from that of Lemma D.4 in \cite{Agarwal:2020_pg_theory}, as the second-order gradients are identical. We then have that $\Omega(\theta)$ is 2-smooth and therefore $\frac{\lambda}{|\St|} \Omega(\theta)$ is $\frac{2\lambda}{|\St|}$-smooth, completing the proof. 
\end{proof}

Note that the second value of $\lambda$ will nearly always be greater than $1$ for most values of $\eps, \epsilon_{\mathrm{opt}},|\St|, |\A|$, as that's the case when $\expect[\mu]{\alpha(s)} > \frac{(1 - \gamma)^2\eps}{|\A| - 1} - \epsilon_{\mathrm{opt}}|\St|$, which is usually negative, thus trivially satisfying the inequality for $\alpha(s)\in[0,1]$ $\forall s\in\St$.

\itercomplexity*

\begin{proof}
    The proof rests on bounding the iteration complexity of making the gradient sufficiently small. Because the optimization process is deterministic and unconstrained, we can use the standard result that after $T$ updates with stepsize $1/\beta_\lambda$, we have 
    \begin{align}
        \min_{t\leq T} ||\grad_\theta \J(\theta^{(t)})||_2^2 \leq \frac{2\beta_\lambda 
        (\J(\theta^*) - \J(\theta^{(0)}))}{T} = \frac{2\beta_\lambda}{(1-\gamma)T},
    \end{align}
    where $\beta_\lambda$ upper-bounds the smoothness of $\J(\theta)$. Using the above and \cref{thm:alpha_reg_sd}, we want 
    \begin{align*}
        \eps_{\mathrm{opt}} \leq \sqrt{\frac{2\beta_\lambda}{(1-\gamma)T}} \leq \frac{\lambda}{2|\St||\A|}. 
    \end{align*}
    Solving the above inequality for $T$ gives $T \geq \frac{8|\St|^2|\A|^2\beta_\lambda}{\lambda^2(1-\gamma)}$. From \cref{thm:smoothness_jalpha}, we can set $\beta_\lambda = \frac{8}{(1 - \gamma)^3} + \frac{2\lambda}{|\St|}$. Plugging this in gives 
    \begin{align*}
     T\geq   \frac{8|\St|^2|\A|^2\beta_\lambda}{(1-\gamma)\lambda^2} &= \left(\frac{64|\St|^2|\A|^2}{(1 - \gamma)^4\lambda^2} + \frac{16|\St||\A|^2}{(1-\gamma)\lambda}\right).
    \end{align*}
    \cref{thm:alpha_reg_sd} gives us the possible values for $\lambda$ for value error margin $\eps$. Then if 
    \begin{align*}
        \lambda = \frac{\eps (1 - \gamma)}{\expect[\mu]{\kappaA} \Dinf} < 1,
    \end{align*}
    we can write  
    \begin{align*}
        \frac{64|\St|^2|\A|^2}{(1 - \gamma)^4\lambda^2} + \frac{16|\St||\A|^2}{(1-\gamma)\lambda}
        &\leq \frac{80|\St|^2|\A|^2}{(1 - \gamma)^4\lambda^2} \\
        &= \frac{80\expect[\mu]{\kappaA}^2|\St|^2|\A|^2}{\eps^2(1 - \gamma)^6} \Big\| \frac{d_\rho^{\pi^*}}{\mu}\Big\|_\infty^2.
    \end{align*}
\end{proof}

\itercompinit*
\begin{proof}
    The proof rests on bounding the iteration complexity of making the gradient sufficiently small. Because the optimization process is deterministic and unconstrained, we can use the standard result that after $T$ updates with stepsize $1/\beta_\lambda$, we have 
    \begin{align}
        \min_{t\leq T} ||\grad_\theta \J(\theta^{(t)})||_2^2 \leq \frac{2\beta_\lambda 
        (\J(\theta^*) - \J(\theta^{(0)}))}{T} = \frac{2\beta_\lambda}{T}\Delta,
    \end{align}
    where $\beta_\lambda$ upper-bounds the smoothness of $\J(\theta)$ and we. define $\Delta:= \J(\theta^*) - \J(\theta^{(0)}))$ for conciseness. Using the above and \cref{thm:alpha_reg_sd}, we want 
    \begin{align*}
        \eps_{\mathrm{opt}} \leq \sqrt{\frac{2\beta_\lambda\Delta}{T}} \leq \frac{\lambda}{2|\St||\A|}. 
    \end{align*}
    Solving the above inequality for $T$ gives $T \geq \Delta\frac{8|\St|^2|\A|^2\beta_\lambda}{\lambda^2}$. From \cref{thm:smoothness_jalpha}, we can set $\beta_\lambda = \frac{8}{(1 - \gamma)^3} + \frac{2\lambda}{|\St|}$. Plugging this in gives 
    \begin{align*}
     T\geq   \Delta\frac{8|\St|^2|\A|^2\beta_\lambda}{\lambda^2} &= \Delta\left(\frac{64|\St|^2|\A|^2}{(1 - \gamma)^3\lambda^2} + \frac{16|\St||\A|^2}{\lambda}\right).
    \end{align*}
    \cref{thm:alpha_reg_sd} ensures that $\min_{t\leq T} V^{\star}(\rho)-V^{(t)}(\rho)\leq \epsilon$ provided that $\lambda$ is of the form:
    \begin{align*}
        \lambda = \frac{\eps (1 - \gamma)}{\expect[\mu]{\kappaA} \Dinf} < 1,
    \end{align*}
    we can therefore write:  
    \begin{align*}
        \frac{64|\St|^2|\A|^2}{(1 - \gamma)^3\lambda^2} + \frac{16|\St||\A|^2}{\lambda}
        &\leq \frac{80|\St|^2|\A|^2}{(1 - \gamma)^3\lambda^2} \\
        &= \frac{80\expect[\mu]{\kappaA}^2|\St|^2|\A|^2}{\eps^2(1 - \gamma)^5} \Big\| \frac{d_\rho^{\pi^*}}{\mu}\Big\|_\infty^2.
    \end{align*}
This implies the following condition on $T$:
\begin{align}
    T\geq \frac{80\Delta\vert\mathcal{A} \vert^2 \vert \mathcal{S}\vert^2}{\epsilon^2(1-\gamma)^5}\mathbb{E}_{\mu} \left[\kappa^{\alpha}_{\mathcal{A}}(s)\right]^2  \left\Vert\frac{d_{\rho}^{\pi^{\star}}}{\mu} \right\Vert_{\infty}^2
\end{align}
It remains to control the error $\Delta$ due to initialization with policy $\pi_0$.
Denote by $\pi_{\lambda}^{\star}$ the optimal policy maximizing $\mathcal{J}_{\lambda}^{\star}$.  We have the following:
\begin{align}
\begin{split}
    \Delta := & V^{\pi_{\lambda}^{\star}}(\rho)-V^{\pi_{0}}(\rho) - \lambda \kl(\pi_0, \pi_{\lambda}^{\star})\\
    \leq &
    V^{\pi_{\lambda}^{\star}}(\rho)-V^{\star}(\rho) + V^{\star}(\rho)-V^{\pi_{0}}(\rho)\\
    \leq & V^{\star}(\rho)-V^{\pi_{0}}(\rho)\\
    \leq &
    \frac{1}{(1-\gamma)^2} \left\Vert \frac{d_{\rho}^{\pi_0}}{\mu'} \right\Vert_{\infty}\mathbb{E}_{s\sim \mu'}\left[\alpha(s)\right] 
\end{split}
\end{align}
where the first line is by definition of $\Delta$, the second line uses that the KL term is non-positive. The third line uses that $ V^{\pi_{\lambda}^{\star}}(\rho)-V^{\star}(\rho)\leq 0$ and the last line follows from  \cref{lemm:lower_bound_value}.  Hence, it suffice to choose $T$ satisfying:
\begin{align}
    T\geq \frac{80\vert\mathcal{A} \vert^2 \vert \mathcal{S}\vert^2}{\epsilon^2(1-\gamma)^7} \left\Vert\frac{d_{\rho}^{\pi^{\star}}}{\mu} \right\Vert_{\infty}^2\left\Vert \frac{d_{\rho}^{\pi_0}}{\mu'} \right\Vert_{\infty}\mathbb{E}_{\mu} \left[\kappa^{\alpha}_{\mathcal{A}}(s)\right]^2 \mathbb{E}_{s\sim \mu'}\left[\alpha(s)\right]
\end{align}
As a final step, we simply observe that $\mathbb{E}_{\mu} \left[\kappa^{\alpha}_{\mathcal{A}}(s)\right] \leq 2$ and $d_\rho^{\pi_0} \leq 1$.

\end{proof}

\begin{lemma}
    Following \cref{thm:smoothness_jalpha}, let $\beta_\lambda= \frac{8\gamma}{(1-\gamma)^3} + \frac{2\lambda}{|\St|}$. From any initial $\theta^{(0)}$ and following \cref{eq:grad_ascent} with $\eta = 1/\beta_\lambda$ and 
    \begin{align}
        \lambda = 
        \frac{\eps_{\mathrm{opt}}|\St|(|\A| - 1)}{(1 - \gamma)^2 \eps - (|\A| - 1)\expect[\mu]{\alpha(s)}},  
    \end{align}
    for all starting state distributions $\rho$, we have,
    \begin{align}
    \begin{split}
        \min_{t<T} \{V^*(\rho) &- V^{(t)}(\rho)\} \leq \eps  \\
        \textit{whenever} 
        \quad
        T &\geq  \min\left\{\frac{80\expect[\mu]{\kappaA}^2|\St|^2|\A|^2}{(1 - \gamma)^6 \eps^2} \Big\| \frac{d_\rho^{\pi^*}}{\mu}\Big\|_\infty^2,
        \ 80|\St||\A|^2\left(\frac{\eps}{\epsilon_{\mathrm{opt}}(1 - \gamma)^2(|\A| - 1)} - \frac{\expect[\mu]{\alpha(s)}}{\epsilon_{\mathrm{opt}}(1-\gamma)^4} \right)
        \right\}.
    \end{split}
    \end{align}
   
\end{lemma}

\begin{proof}
    The proof proceeds identically as above, except we set
    \begin{align*}
        \lambda = \frac{\epsilon_{\mathrm{opt}}|\St|(|\A| - 1)}{(1 - \gamma)^2 \eps - (|\A| - 1)\expect[\mu]{\alpha(s)}} > 1,
    \end{align*}
    we have 
    \begin{align*}
        \frac{64|\St|^2|\A|^2}{(1 - \gamma)^4\lambda^2} + \frac{16|\St||\A|^2}{(1-\gamma)\lambda}
            &\leq \frac{80|\St|^2|\A|^2}{(1 - \gamma)^4\lambda} \\
            &= \frac{80|\St||\A|^2\left((1 - \gamma)^2 \eps - (|\A| - 1)\expect[\mu]{\alpha(s)}\right)}{(1 - \gamma)^4 \eps_{\mathrm{opt}}(|\A - 1)} \\
            &= 80|\St||\A|^2\left(\frac{\eps}{\epsilon_{\mathrm{opt}}(1 - \gamma)^2(|\A| - 1)} - \frac{\expect[\mu]{\alpha(s)}}{\epsilon_{\mathrm{opt}}(1-\gamma)^4} \right)
    \end{align*}
    completing the proof. 
\end{proof}
Note that this value of $\lambda$ will nearly always be greater than $1$ for most values of $\eps, \epsilon_{\mathrm{opt}},|\St|, |\A|$, as that's the case when $\expect[\mu]{\alpha(s)} > \frac{(1 - \gamma)^2\eps}{|\A| - 1} - \epsilon_{\mathrm{opt}}|\St|$, which is usually negative, thus trivially satisfying the inequality for $\alpha(s)\in[0,1]$ $\forall s\in\St$.

\begin{lemma}\label{lemm:lower_bound_value}
Assume that $\pi$ is such that $\pi(a^{\star}(s)|s)\geq 1- \beta(s)$ for some state dependent error $s\mapsto \beta(s)$ and that $\rho(s)>0$ for all states $s$. Then the following inequality holds:
\begin{align}
    V^{\pi}(\rho) -V^*(\rho)\geq -\frac{1}{(1-\gamma)^2} \left\Vert \frac{d_{\rho}^{\pi}}{\rho} \right\Vert_{\infty} \mathbb{E}_{\rho}\left[ \beta(s) \right]
\end{align}

\end{lemma}
\begin{proof}

     \begin{align*}
      V^{\pi}(\rho) -V^*(\rho) &= \frac{1}{1-\gamma} \sum_{s,a} d_{\rho}^{\pi}(s) \pi(a|s) A^{\pi^*}(s,a) \\
      &=  \frac{1}{1-\gamma} \sum_s \sum_{a\neq a^{\star}(s)}\left( d_{\rho}^{\pi}(s) \pi(a|s) A^{\pi^*}(s,a)   \right) \\
     &\geq  -\frac{1}{(1-\gamma)^2} \sum_s d_{\rho}^{\pi}(s)\sum_{a\neq a^{\star}(s)}    \pi(a|s)   \\
     & \geq 
     -\frac{1}{(1-\gamma)^2} \sum_s \left(  d_{\rho}^{\pi}(s) \beta(s)   \right)\\
     &\geq 
     -\frac{ 1}{(1-\gamma)^2} \left\Vert \frac{d_{\rho}^{\pi}}{\mu} \right\Vert_{\infty}\mathbb{E}_{s\sim \mu}\left[\beta(s)\right] 
 \end{align*}
where the first line follows by application of the performance different lemma \cite[Lemma 3.2]{Agarwal:2020_pg_theory},  the second line is  due to the fact that $A^{\pi^*}(s,a^*(s)) = 0$, the third line from $A^{\pi^*}(s,a)\geq -1/(1-\gamma) $  for $a\neq a^*$. The fourth line uses that $\sum_{a\neq a^{\star}(s)}\pi(a|s) = 1-\pi(a^{\star}(s)|s) \leq \beta(s) $ for $a\neq a^{\star}(s)$ since by assumption $\pi(a^{\star}(s)|s)\geq 1- \beta(s)$. Finally, the last line uses that $d_{\rho}^{\pi}$ is a probability distribution over states $s$ satisfying $\sum_{s\in \mathcal{S}} d_{\rho}^{\pi}(s)=1$.

\end{proof}

\subsection{State dependent $\lambda$  and $\epsilon$}

We can further generalize these results by allowing the error $\epsilon$ and regularization weight $\lambda$ to be state-dependent. The gradient with state dependent regularized $\lambda$ equals

\begin{align*}
        \mathcal{J}^{\pi_0}(\theta) = V^{\pi_\theta}(\mu) + \sum_{s,a}  \frac{\lambda(s)}{|\St|}\pi_0(a|s) \log \pitheta  
    \end{align*}

\begin{lemma} \label{thm:alpha_reg_sd_state_dependent}
    Suppose $\theta$ is such that $\left( \nabla \Jz(\theta) \right)_{s,a} \leq \epsilon_{\mathrm{opt}}(s,a)$. Then we have that for all states $s\in\St$,

      \begin{align*}
        V^{\pi_\theta}(\rho) &\geq V^*(\rho) - \min\Biggr\{
        \frac{1}{1-\gamma} \expect[s\sim\mathrm{Unif}_{\St}]{\frac{\epsilon_{\mathrm{opt}}(s,a^*(s)) |\mathcal{S}|}{\max\left((1 - \alpha(s)) - \frac{\epsilon_{\mathrm{opt}}(s,a^*(s))|\mathcal{S}|}{\lambda(s)}, 0\right)}  + \lambda(s) \alpha(s)} \Dinf,\\
        &\quad  \frac{|\A| }{(1-\gamma)^2}\mathbb{E}_{s \sim \mu   }  \left[ \alpha(s) \right] \left\| \frac{d_{\rho}^{\pi_\theta}}{\mu}\right\|_\infty+  \frac{|\St|}{(1-\gamma)^2} \left\| \frac{\sum_{a} \epsilon_{\mathrm{opt}}(s,a) }{\lambda(s)} \right\|_\infty
        \Biggr\}
    \end{align*}

\end{lemma}

\begin{proof}
    Let's assume that $\pi^*$ is a deterministic policy. By \cite{Puterman:2010} such an optimal policy always exists for an MDP. We'll use the notation $a^*(s)$ to denote the optimal action at state $s$. This, combined with the assumption that $d_{\mathrm{TV}}(\pi^*(\cdot|s),\pi_0(\cdot|s)) \leq \alpha(s)$ for all $s\in\St$, tells us that $\pi_0(a^*(s)|s) \geq \pi^*(a^*(s)|s) - \alpha(s) = 1 - \alpha(s)$. Similarly, for $a\neq a^*(s)$, $\pi_0(a|s) \leq \alpha(s)$. Using this, we can start by showing that whenever $A^{\pi_\theta}(s, a^*(s)) \geq 0$ we can lower bound $\pi_\theta(a^*(s) | s)$ for all $s$.

            The gradient norm assumption $ \left( \nabla \Jz(\theta) \right)_{s,a} \leq \epsilon_{\mathrm{opt}}(s,a)$ implies that for all $s,a$:
            \begin{equation*}
                \epsilon_{\mathrm{opt}}(s,a) \geq \Partial{\Jz(\theta)}{\theta_{s,a}} = \frac{1}{1 - \gamma} d_\mu^{\pi_\theta}(s) \pitheta A^{\pi_\theta}(s,a) + \frac{\lambda(s)}{|\St|} (\piz - \pitheta)
            \end{equation*}
    
            In particular for all $s$,
            \begin{align}
            \begin{split} \label{eq:grad_sd}
                \epsilon_{\mathrm{opt}}(s,a^*(s)) &\geq \Partial{\Jz(\theta)}{\theta_{s,a^*(s)}} \stackrel{(i)}{\geq} \frac{1}{1 - \gamma} d_\mu^{\pi_\theta}(s) \pi_\theta(a^*(s)|s) A^{\pi_\theta}(s,a^*(s)) + \frac{\lambda(s)}{|\St|} (\pi^*(a^*(s)|s) - \alpha(s) - \pi_\theta(a^*(s)|s)) \\
                &= \frac{1}{1 - \gamma} d_\mu^{\pi_\theta}(s) \pi_\theta(a^*(s)|s) A^{\pi_\theta}(s,a^*(s)) + \frac{\lambda(s)}{|\St|} (1 - \alpha(s) - \pi_\theta(a^*(s)|s))
            \end{split}
            \end{align}

            And therefore if $A^{\pi_\theta}(s, a^*(s)) \geq 0$,
            \begin{align*}
                \epsilon_{\mathrm{opt}}(s,a) &\geq  \frac{\lambda(s)}{|\St|} (1 - \alpha(s) - \pi_\theta(a^*(s)|s))
            \end{align*}
            Thus,
            \begin{equation}\label{eq:policies_closeness_alpha_sd}
                \pi_\theta(a^*(s)|s) \geq \max\left( 1 - \alpha(s) - \frac{\epsilon_{\mathrm{opt}}(s,a^*(s))|\St|}{\lambda(s)} , 0\right) \geq 1 - \alpha(s) - \frac{\epsilon_{\mathrm{opt}}(s,a^*(s))|\St|}{\lambda(s)}.
            \end{equation}
            In this case we can upper bound $A^{\pi_\theta}(s, a^*(s))$. From \cref{eq:grad} inequality $(i)$, we have 
            \begin{align*}
                A^{\pi_\theta}(s, a^*(s)) &\leq \frac{1-\gamma}{d_\mu^{\pi_\theta}(s)}\left(\frac{1}{\pi_\theta(a^*(s)|s)} \Partial{\Jz(\theta)}{\theta_{s,a^*(s)}} - \frac{\lambda(s)}{|\St|} \frac{1}{\pi_\theta(a^*(s)|s)} \left(1 - \alpha(s) - \pi_\theta(a^*(s)|s)\right) \right) \\ 
                &= \frac{1-\gamma}{d_\mu^{\pi_\theta}(s)}\left(\frac{1}{\pi_\theta(a^*(s)|s)} \Partial{\Jz(\theta)}{\theta_{s,a^*(s)}} + \frac{\lambda(s)}{|\St|}\left(1 - \frac{1 - \alpha(s)}{\pi_\theta(a^*(s)|s)} \right) \right) \\
                &\stackrel{(i)}{\leq} \frac{1}{\mu(s)}\left(\frac{1}{\max\left((1 - \alpha(s)) - \frac{\epsilon_{\mathrm{opt}}(s,a^*(s))|\mathcal{S}|}{\lambda}, 0\right)} \cdot \epsilon_{\mathrm{opt}}(s,a^*(s)) + \frac{\lambda(s)}{|\St|}\left( 1 - (1 - \alpha(s))\right) \right) \\
                &\leq \frac{1}{\mu(s)}\left(\frac{1}{\max\left((1 - \alpha(s)) - \frac{\epsilon_{\mathrm{opt}}(s,a^*(s))|\mathcal{S}|}{\lambda}, 0\right)} \cdot \epsilon_{\mathrm{opt}}(s,a^*(s))+ \frac{\lambda(s)}{|\St|} \alpha(s)\right) 
            \end{align*}

            Where $(i)$ follows because $d_{\mu}^{\pi_\theta}(s)\geq (1-\gamma)\mu(s)$, $\frac{\partial \mathcal{J}_\lambda^{\alpha}(\theta)}{\partial \theta_{s,a}}\leq \epsilon_{\mathrm{opt}}$ and $\max\left(1 - \alpha(s) - \frac{\epsilon_{\mathrm{opt}}(s, a^*(s))|\St|}{\lambda}, 0\right) \leq \pi_\theta(a^*(s) | s) \leq 1$ . 
            
           
            We now make use of the performance difference lemma:
           
 \begin{align*}
             V^*(\rho) - V^{\pi_\theta}(\rho) &= \frac{1}{1-\gamma} \sum_{s,a} d_{\rho}^{\pi^*}(s) \pi^*(a|s) A^{\pi_\theta}(s,a) \\
                &= \frac{1}{1-\gamma} \sum_{s} d_{\rho}^{\pi^*}(s)  A^{\pi_\theta}(s,a^*(s)) \\
                &\leq \frac{1}{1-\gamma} \sum_{s} d_{\rho}^{\pi^*}(s)  A^{\pi_\theta}(s,a^*(s)) \mathbf{1}( A^{\pi_\theta}(s,a^*(s)) \geq 0 ) \\
                &\leq \frac{1}{1-\gamma}  \sum_{s} \frac{d_{\rho}^{\pi^*}(s) }{\mu(s)} \left( \frac{1}{\max\left((1 - \alpha(s)) - \frac{\epsilon_{\mathrm{opt}}(s,a)|\mathcal{S}|}{\lambda}, 0\right)} \cdot \epsilon_{\mathrm{opt}}(s,a^*(s)) + \frac{\lambda(s)}{|\St|} \alpha(s) \right)\\
                &\qquad\qquad\qquad\qquad\qquad\qquad\qquad\qquad\times\mathbf{1}( A^{\pi_\theta}(s,a^*(s)) \geq 0 )  \\ 
                &\leq \frac{1}{1-\gamma} \expect[s\sim\mathrm{Unif}_{\St}]{\frac{\epsilon_{\mathrm{opt}}(s,a^*(s)) |\mathcal{S}|}{\max\left((1 - \alpha(s)) - \frac{\epsilon_{\mathrm{opt}}(s,a^*(s)|\mathcal{S}|}{\lambda(s)}, 0\right)}  + \lambda(s) \alpha(s)} \Dinf
         \end{align*}


 Now let's relate the values of $\pi^*$ and $\pi_\theta$. We will again apply the performance difference lemma, this time in the other direction:
     \begin{align*}
      V^{\pi_\theta}(\rho) -V^*(\rho) &= \frac{1}{1-\gamma} \sum_{s,a} d_{\rho}^{\pi_\theta}(s) \pi_\theta(a|s) A^{\pi^*}(s,a) \\
      &\eqtext{(1)}  \frac{1}{1-\gamma} \sum_s \left( \sum_{a\neq a^*(s)} d_{\rho}^{\pi_\theta}(s) \pi_\theta(a|s) A^{\pi^*}(s,a)   \right) \\
      &\stackrel{(2)}{\geq }  \frac{-1}{1-\gamma} \sum_s \sum_a d_{\rho}^{\pi_\theta}(s) \left(\alpha(s)(|\mathcal{A}|-1) + \frac{\sum_{a\neq a^*(s)} \epsilon_{\mathrm{opt}}(s,a)|\St|}{\lambda(s)}\right) \frac{1}{1-\gamma} \\
      &= -\frac{ 1}{1-\gamma} \left(\sum_s \frac{d_{\rho}^{\pi_\theta}(s)}{1 - \gamma}\alpha(s)(|\A| -1) + \sum_s |\St|d_{\rho}^{\pi_\theta}(s) \sum_{a\neq a^*(s)}\frac{\epsilon_{\mathrm{opt}}(s,a)}{(1 - \gamma )\lambda(s)}\right) \quad \\ 
    &\stackrel{(3)}{\geq} - \sum_s \frac{d_{\rho}^{\pi_\theta}(s)}{(1 - \gamma)^2}\alpha(s)(|\A|-1) - \frac{|\St|}{(1-\gamma)^2} \left\| \frac{\sum_{a \neq a^*(s)} \epsilon_{\mathrm{opt}}(s,a) }{\lambda(s)} \right\|_\infty   \\ 
      &\geq  -\frac{|\A|}{(1-\gamma)^2}\mathbb{E}_{s \sim d_{\rho}^{\pi_\theta}  }  \left[ \alpha(s) \right] -   \frac{|\St|}{(1-\gamma)^2} \left\| \frac{\sum_{a} \epsilon_{\mathrm{opt}}(s,a) }{\lambda(s)} \right\|_\infty   \\
      &\geq -\frac{|\A| }{(1-\gamma)^2}\mathbb{E}_{s \sim \mu   }  \left[ \alpha(s) \right] \left\| \frac{d_{\rho}^{\pi_\theta}}{\mu}\right\|_\infty-   \frac{|\St|}{(1-\gamma)^2} \left\| \frac{\sum_{a} \epsilon_{\mathrm{opt}}(s,a) }{\lambda(s)} \right\|_\infty 
 \end{align*}
where (1) is due to the fact that $A^{\pi^*}(s,a^*(s)) = 0$, and (2) is due to the fact that $A^{\pi^*}(s,a)$ for $a\neq a^*$ is lower-bounded by $-1/(1-\gamma)$ and \cref{eq:policies_closeness_alpha_sd}. and $(3)$ holds because of Holder and $\sum_{s} d_{\rho}^{\pi_\theta}(s) = 1$. 
Therefore,

\begin{equation*}
    V^{\pi_\theta}(\rho) + \frac{|\A| }{(1-\gamma)^2}\mathbb{E}_{s \sim \mu   }  \left[ \alpha(s) \right] \left\| \frac{d_{\rho}^{\pi_\theta}}{\mu}\right\|_\infty+  \frac{|\St|}{(1-\gamma)^2} \left\| \frac{\sum_{a} \epsilon_{\mathrm{opt}}(s,a) }{\lambda(s)} \right\|_\infty  \geq V^*(\rho)  .
\end{equation*}

\end{proof}

A simple corollary of Lemma~\ref{thm:alpha_reg_sd_state_dependent} is,
\begin{corollary}\label{corollary::state_dependent_alpha_epsilon}
If $\epsilon_{\mathrm{opt}}(s,a) \leq \frac{(1-\alpha(s))\lambda(s)}{|\St|}$ then
\begin{align*}
        V^{\pi_\theta}(\rho) &\geq V^*(\rho) - \min\Biggr\{
        \frac{1}{1-\gamma} \expect[s\sim\mathrm{Unif}_{\St}]{\frac{2\epsilon_{\mathrm{opt}}(s,a^*(s)) |\mathcal{S}|}{1-\alpha(s)}  + \lambda(s) \alpha(s)} \Dinf,\\
        &\quad  \frac{|\A| }{(1-\gamma)^2}\mathbb{E}_{s \sim \mu   }  \left[ \alpha(s) \right] \left\| \frac{d_{\rho}^{\pi_\theta}}{\mu}\right\|_\infty+  \frac{|\St|}{(1-\gamma)^2} \left\| \frac{\sum_{a} \epsilon_{\mathrm{opt}}(s,a) }{\lambda(s)} \right\|_\infty
        \Biggr\}
    \end{align*}
\end{corollary}
\begin{proof}
If $\epsilon(s,a) \leq \frac{(1-\alpha(s))\lambda(s)}{|\St|}$ then $\max\left((1 - \alpha(s)) - \frac{\epsilon_{\mathrm{opt}}(s,a)|\mathcal{S}|}{\lambda(s)}, 0\right) \geq \frac{1-\alpha(s)}{2}$. The result follows.
\end{proof}

Corollary~\ref{corollary::state_dependent_alpha_epsilon} recovers the results of~\cite{Agarwal:2020_pg_theory} by noting the TV distance between the optimal policy and the uniform one equals $1- \frac{1}{|\A|}$ and therefore $1-\alpha(s) = \frac{1}{|\A|}$.

We now concern ourselves with the problem of finding a \underline{true} $\epsilon > 0$ optimal policy. This will require us to set the values of $\lambda(s)$ appropriately. We restrict ourselves to the following version of the results of Corollary~\ref{corollary::state_dependent_alpha_epsilon}. If $\epsilon(s,a) \leq \frac{(1-\alpha(s))\lambda(s)}{|\St|}$ then

\begin{align*}
     V^{\pi_\theta}(\rho) &\geq V^*(\rho) - 
        \frac{1}{1-\gamma} \expect[s\sim\mathrm{Unif}_{\St}]{\frac{2\epsilon_{\mathrm{opt}}(s,a^*(s)) |\mathcal{S}|}{1-\alpha(s)}  + \lambda(s) \alpha(s)} \Dinf
\end{align*}

By setting $\lambda(s) = \frac{\epsilon(1-\gamma)}{2\alpha(s)\Dinf}$ and $\epsilon_{\mathrm{opt}}(s,a) = \min\left(\frac{(1-\alpha(s)) \epsilon (1-\gamma) }{4|\St|\Dinf} , \frac{(1-\alpha(s))\lambda(s)}{|\St|}\right)$ we get
\begin{equation*}
      V^{\pi_\theta}(\rho) \geq V^*(\rho) -  \epsilon.
\end{equation*}

Observe that the level of regularization depends on the state's error. If the error is very low, the regularizer $\lambda(s)$ should be set to a larger value.



\section{Multitask learning} \label{sec:appendix_multitask}

Assume we are given $K$ i.i.d. tasks $M_k$ sampled from $\mathcal P_\mathcal{M}$, denote by $\pi_k^{\star}(\cdot|s)$ their corresponding optimal policies and let $\tilde{\pi}_k(\cdot|s)$ be $\alpha(s)$ policies, i.e.  $d_{TV}(\tilde\pi_k(\cdot|s),\pi_k^{\star}(\cdot|s))\leq \alpha(s)$ for some $\alpha(s)\leq 1$. To simplify notation, we may also refer to $\mathcal{P}$ directly as the distribution over these optimal policies. Let  $\hat{\pi}_0$ be the total variation barycenter of the policies $\tilde{\pi}_k$, i.e.: $\hat{\pi}_0 = \arg\min_{\pi} \frac{1}{K} \sum_{k=1}^K d_{TV}(\pi,\tilde{\pi}_i) $,  while $\pi_0 = \arg\min_{\pi} \mathbb{E}_{M_k\sim \mathcal P_\mathcal{M}}[ d_{TV}(\pi,\pi_i^{\star})]$. 

\tvbarycenter*
\begin{proof}
Let's first express the barycenter loss in a more convenient form:
\begin{align}
    \mathbb{E}_{\pi'\sim P}\left[ d_{TV}(\pi(.|s),\pi'(.|s)) \right] = &\mathbb{E}_{\pi'\sim P}\left[ \frac{1}{2} \sum_{a\neq a_{\pi'}(s)} \pi(a) +\frac{1}{2}(1-\pi(a_{\pi'}(s),s))\right]\\
    =& 
    \mathbb{E}_{\pi'\sim P}\left[(1-\pi(a_{\pi'}(s),s))\right]\\
    =& 1-\sum_{a} \mathbb{P}(\pi'(a|s)=1)\pi(a|s)\\
    =& 1-\sum_{a} \pi_{soft}(a|s)\pi(a|s).
\end{align}
Therefore, the barycenter loss is minimized when $\pi(a|s)$  puts all its mass on the maximum value of $\pi_{soft}(a|s)$ over actions $a\in \mathcal{A}$.

\end{proof}

The KL barycenter can be described as follows. 
\begin{lemma}[KL barycenter] \label{thm:kl_bary}
    Let $\mathcal{P}_\mathcal M$ be a distribution over tasks such that for every $M_k \in \mathcal M$, there exists a unique optimal action $a^\star_k(s)$ for each state $s$ such that $\pi_k^\star(s)=a^\star$. Then the KL barycenter for state $s$ is:
    \begin{align}
        \argmin_\pi \mathbb E_{M_k\sim\mathcal P_\mathcal M} \kl(\pi_k^\star(\cdot|s), \pi(\cdot|s)) = \delta(a = \mathbb E_{M_k\sim\mathcal P_\mathcal M} \pi^\star_k(s))
    \end{align}
    where $\delta(\cdot)$ is the Dirac delta distribution. This holds for both directions of the KL. 
    \end{lemma}
    \begin{proof}
    We have 
    \begin{align*}
        \mathbb E_{M_k\sim\mathcal P_\mathcal M} \kl(\pi_k^\star(\cdot|s), \pi(\cdot|s)) &=  
        \mathbb E_{M_k\sim\mathcal P_\mathcal M} \sum_a \pi_k^\star(a|s) \log \frac{\pi_k^\star(a|s)}{\pi(a|s)} \\
        &= \mathbb E_{M_k\sim\mathcal P_\mathcal M} \left[-\log \pi(a^\star_k(s)|s) + \underbrace{\sum_{a\neq a^\star_k(s)} 0 \cdot \log \frac{0}{\pi(a|s)}}_{=0}\right]\\
        &= \mathbb E_{M_k\sim\mathcal P_\mathcal M} [-\log \pi(a^\star_k(s)|s)]
    \end{align*}
    Therefore, the barycenter loss is minimized when $\pi(a|s)$  puts all its mass on the expected $a^\star_k(s)$. Note that we consider the underbrace term zero because $\lim_{x\to0} x\log x = 0$. It is easy to verify that this result holds for the reverse KL. 
    \end{proof}



\multitaskiters*
\begin{proof}
Let $M_i$ be a random task sampled according to $\mathcal{M}$ and denote by $\pi_i^{\star}$ its corresponding optimal policy. Set $\alpha(s) = d_{TV}(\hat{\pi}_0,\pi_i^{\star})$ and choose $\lambda = \frac{\epsilon(1-\gamma)}{2\Vert\frac{d_{\rho}^{\pi^{\star}}}{\mu} \Vert } $. 
By \cref{thm:jalpha_iters}, we have that:
    \begin{align}
    \begin{split}
        \min_{t<T} \{V^*(\rho) &- V^{(t)}(\rho)\} \leq \eps  \\
        \textit{whenever} 
        \quad
        T &\geq  \frac{160\vert\mathcal{A} \vert^2 \vert \mathcal{S}\vert^2}{\epsilon^2(1-\gamma)^7} \left\Vert\frac{d_{\rho}^{\pi^{\star}}}{\mu} \right\Vert_{\infty}^2\left\Vert \frac{d_{\rho}^{\hat{\pi}_0}}{\mu'} \right\Vert_{\infty} \mathbb{E}_{s\sim \mu'}\left[\alpha(s)\right].
    \end{split}
    \end{align}
By choosing $\mu'$ to be uniform and recalling that $d_{\rho}^{\hat{\pi}_0}\leq 1$, it suffice to have:
\begin{align}
     T &\geq  \frac{160\vert\mathcal{A} \vert^2 \vert \mathcal{S}\vert^3}{\epsilon^2(1-\gamma)^7} \left\Vert\frac{d_{\rho}^{\pi^{\star}}}{\mu} \right\Vert_{\infty}^2 \mathbb{E}_{s\sim \mu'}\left[d_{TV}(\hat{\pi}_0,\pi_i^{\star})\right].
\end{align}
Taking the expectation over the tasks and treating $T$ as a random variable depending on the task, we get that:
\begin{align}
    \mathbb{E}\left[T\right]\geq  \frac{160\vert\mathcal{A} \vert^2 \vert \mathcal{S}\vert^3}{\epsilon^2(1-\gamma)^7} \left\Vert\frac{d_{\rho}^{\pi^{\star}}}{\mu} \right\Vert_{\infty}^2 \mathbb{E}_{s\sim \mu'}\mathbb{\pi'\sim P}\left[d_{TV}(\hat{\pi}_0,\pi')\right].
\end{align}

\end{proof}

  The following lemma quantifies how  $\hat{\pi}_0$ is close  to be the TV barycenter of $\{\pi_k^{\star}\}_{1\leq k\leq K}$ when $K$ grows to infinity. We let $\tilde{\pi}_k(\cdot|s)$ be, on average, $\zeta(s)$-optimal in state $s$ across tasks $M_k$, i.e.  $\mathbb{E}_{M_k\sim \mathcal{M}}\left[\tv(\tilde\pi_k(\cdot|s),\pi_k^{\star}(\cdot|s))\right]\leq \zeta(s)$ for some $\zeta(s)\in [0, 1]$. For concision, we shorten $\pi(\cdot|s)$ as $\pi$.

\barycentercon*
\begin{proof}
To simplify the proof, we fix a state $s$ and omit the dependence in $s$. We further introduce the following notations:
\begin{align}
    f(\pi) &= \mathbb{E}_{M_i\sim \mathcal P_\mathcal{M}}\left[ d_{TV}(\pi,\pi_i^{\star}) \right]\\
    \tilde{f}(\pi) &= \frac{1}{K} \sum_{i=1}^K d_{TV}(\pi,\tilde{\pi}_{i})\\
    \hat{f}(\pi) &= \frac{1}{K} \sum_{i=1}^K  d_{TV}(\pi,\pi_{i}^{\star})
\end{align}
Let  $\pi_0 = \arg\min_{\pi} f(\pi)$ and $\hat{\pi}_0=\arg\min_{\pi}\tilde{f}(\pi)$.
It is easy to see that:
\begin{align*}
     f(\hat{\pi}_0)\leq &  \tilde{f}(\hat{\pi}_0) + \vert\hat{f}(\hat{\pi}_0) -f(\hat{\pi}_0)  \vert + \vert \tilde{f}(\hat{\pi}_0)-\hat{f}(\hat{\pi}_0)\vert \\
        \leq & \hat{f}(\pi_0) + \vert\hat{f}(\hat{\pi}_0) -f(\hat{\pi}_0)  \vert + \vert \hat{f}(\hat{\pi}_0)-\hat{f}(\hat{\pi}_0)\vert\\
        \leq &  f(\pi_0) +\vert\hat{f}(\hat{\pi}_0) -f(\hat{\pi}_0)  \vert + \vert \tilde{f}(\hat{\pi}_0)-\hat{f}(\hat{\pi}_0)\vert\\
         +& \vert\hat{f}(\pi_0) -f(\pi_0)  \vert + \vert \tilde{f}(\pi_0)-\hat{f}(\pi_0)\vert\\
        \leq & f(\pi_0) + 2\sup_{\pi} \vert \hat{f}(\pi)-f(\pi) \vert  + 2\sup_{\pi}\vert \hat{f}(\pi)-\tilde{f}(\pi) \vert. 
\end{align*}

where the first line follows by a triangular inequality, the second line uses that $\hat{f}(\hat{\pi}_0)\leq \hat{f}(\pi_0)$ since $\hat{\pi}_0$ minimizes $\hat{f}$. The third line uses a triangular inequality again while the last line follows by definition of the supremum. Moreover, recall that $f(\pi_0)\leq f(\hat{\pi}_0)$ as $\pi_0$ minimizes $f$ and that $\vert \hat{f}(\pi)- \tilde{f}(\pi)\vert \leq \zeta $ since, by assumption, we have that $d_{TV}(\pi_i^{\star},\tilde{\pi}_i)\leq \zeta$. Therefore, it follows that:
\begin{align}\label{eq:main_bound}
    \vert f(\hat{\pi}_0)-f(\pi_0)\vert\leq 2\zeta + 2\sup_{\pi}\vert\hat{f}(\pi)-f(\pi) \vert. 
\end{align}
By application of the bounded difference inequality (McDiarmid's inequality) \cite[Theorem 13.8]{Sen:2018a}, we know that for any $t>0$:
\begin{align}\label{eq:concentration_1}
    \mathbb{P}\left[ \vert \sup_{\pi}\vert\hat{f}(\pi)-f(\pi)\vert  - \mathbb{E}\left[\sup_{\pi}\vert\hat{f}(\pi)-f(\pi)\vert \right] \vert >t  \right]\leq 2e^{-2t^2K}
\end{align}
This implies that for any $0<\eta<1$, we have with probability higher than $1-\eta$ that:
\begin{align}\label{eq:concentration_2}
 \sup_{\pi}\vert\hat{f}(\pi)-f(\pi)\vert \leq \sqrt{\frac{\log(\frac{2}{\delta})}{2K}} + \mathbb{E}\left[\sup_{\pi}\vert\hat{f}(\pi)-f(\pi)\vert \right]
\end{align}
Combining \cref{eq:main_bound} with \cref{eq:concentration_2} and using  \cref{lem:maximal_ineq} to control $\mathbb{E}\left[\sup_{\pi}\vert\hat{f}(\pi)-f(\pi)\vert \right]$, we have that for any $0<\delta<1$, with  probability higher than $1-\delta$, it holds that:
\begin{align}
    \vert f(\hat{\pi}_0)-f(\pi_0) \vert \leq 2\zeta + \sqrt{\frac{2\log(\frac{2}{\delta})}{K}} + 2C\sqrt{\frac{\vert\mathcal{A} \vert}{K}},
\end{align}
for some constant $C$ that depends only on $\vert \mathcal{A}\vert$.

\end{proof}

\begin{lemma}\label{lem:maximal_ineq}

\begin{align}
    \mathbb{E}\left[ \sup_{\pi} \vert \hat{f}(\pi)- f(\pi)\vert   \right]\leq C\sqrt{\frac{\vert \mathcal{A} \vert }{N}},
\end{align}
where $C$ is a constant that depends only on $\vert\mathcal{A} \vert$.
\end{lemma}
\begin{proof}
To control the quantity $\mathbb{E}\left[ \sup_{\pi} \vert \hat{f}(\pi)- f(\pi)\vert   \right]$, we will use a classical result from empirical process theory \cite[Corollary 19.35]{Van-der-Vaart:2000}. We begin by introducing some useful notions to state the result. 
Denote by $\mathcal{F}$ the set of functions $\pi'\mapsto d_{TV}(\pi,\pi')$ that are indexed by a fixed $\pi$. Given a random task $M_i\sim \mathcal{M}$, we call $\pi_i^{\star}$ its optimal policy and denote by $P$ the probability distribution of $\pi_i^{\star}$ when the task $M_i$ is drawn from $\mathcal{M}$.  
Note that we can express $f(\pi)$  as an expectation w.r.t. $P$:  $f(\pi) =  \mathbb{E}_{\pi'\sim P}\left[d_{TV}(\pi,\pi')\right]$. Moreover,  $\hat{f}(\pi)$ is an empirical average over i.i.d. samples $\pi_i^{\star}$ drawn from $P$.

The \emph{bracketing number} $N_{[]}(\epsilon, \mathcal{F},L_2(P))$ is the smallest number of functions $f_j$ and $g_j$ such that for any $\pi$, there exists $j$ such that $ f_j(\pi') \leq d_{TV}(\pi,\pi')\leq g_j(\pi') $ and $\Vert f_j-g_j \Vert_{L_2(P)} \leq \epsilon$. The following result is a direct application of \cite[Corollary 19.35]{Van-der-Vaart:2000} and provides a control on $\mathbb{E}\left[ \sup_{\pi} \vert \hat{f}(\pi)- f(\pi)\vert   \right]$ in terms of the bracketing number $N_{[]}$:
\begin{align}
    \sqrt{N}\mathbb{E}\left[ \sup_{\pi} \vert \hat{f}(\pi)- f(\pi)\vert   \right]\leq \int_0^{R} \sqrt{\log N_{[]}(\epsilon,\mathcal{F}, L_2(P))}.     
\end{align}
where $R^2 =   \mathbb{E}_{\pi'\sim P}\left[ \sup_{\pi} d_{TV}(\pi,\pi')^2 \right] \leq 1$. It remains to control the bracketing number $N_{[]}$.  To achieve this, note that the functions in $\mathcal{F}$ are all $1$-Lipschitz, meaning that:
\begin{align}\label{eq:maximal_ineq}
    \vert d_{TV}(\pi,\pi) - d_{TV}(\pi',\pi)\vert\leq d_{TV}(\pi,\pi')\leq 1. 
\end{align}
Moreover, the family $\mathcal{F}$ admits the constant function $F(\pi')=1$ as an envelope, which means, in other words, that the following upper-bound holds:
\begin{align}
    \sup_{ \pi} d_{TV}(\pi,\pi')\leq 1.
\end{align}
Therefore, we can apply \cite[Example 19.7]{Van-der-Vaart:2000}  to the family $\mathcal{F}$, which directly implies the following  upper-bound on $N_{[]}$:
\begin{align}\label{eq:bracketing_number}
    N_{[]}(\epsilon, \mathcal{F},L_2(P)) \leq K \left({\frac{1}{\epsilon}}\right)^{\vert \mathcal{A} \vert }
\end{align}
where $K$ is a constant that depends only on $\vert\mathcal{A}\vert$. Combining $\ref{eq:maximal_ineq}$ and \ref{eq:bracketing_number} and recalling that $R\leq 1$, it follows that:
\begin{align}
    \mathbb{E}\left[ \sup_{\pi} \vert \hat{f}(\pi)- f(\pi)\vert   \right]\leq C\sqrt{\frac{\vert \mathcal{A} \vert }{N}}.
\end{align}
where $C$ is a constant that depends only on $\vert\mathcal{A} \vert$.
\end{proof}

\section{Experimental details} \label{sec:exp_details} 
The policy model for all algorithms was given by the tabular softmax with single parameter vector $\theta \in \reals^{|\St||\A|}$ such that 
\begin{align*}
    \pi_\theta(a|s) = \frac{\exp(\theta_{s,a})}{\sum_{a'\in\A} \exp(\theta_{s,a'})}. 
\end{align*}
All agents were trained for 80,000 time steps per task using standard stochastic gradient ascent with learning rate $\eta=0.02$. For methods with learned regularizers, the learning for the regularizer was halved, with $\eta_{\mathrm{reg}}=0.01$. Each episode terminated when the agent reached a leaf node. For those using regularization, the regularization weight was $\lambda = 0.2$. For \textsc{Distral}, this weight was applied equally to both the KL term and the entropy term. Each task was randomly sampled with $r(s) = 0$ for all nodes other than the leaf nodes of the subtree rooted at $s_7$ (\cref{fig:tree}). For those nodes, $r(s) \sim \mathrm{Geom}(p)$ with $p=0.5$ for experiments with fixed default policies and $p=0.7$ for those with learned default policies. The sparsity of the reward distribution made learning challenging, and so limiting the size of the effective search space (via an effective default policy) was crucial to consistent success. A single run consisted of 5 draws from the task distribution, with each method trained for 20 runs with different random seeds. For TVPO, the softmax temperature decayed as $\beta(k) = \exp(-k/10)$, with $k$ being the number of tasks. The plotted default policies in \cref{fig:simplex5} were the average default policy probabilities in the selected states across these runs.




\end{document}